\PassOptionsToPackage{dvipsnames}{xcolor}
\documentclass{article}

\usepackage{conference_style,times}

\usepackage{amsmath,amsfonts,bm}

\def\eqref#1{equation~\ref{#1}}

\def\1{\bm{1}}

\DeclareMathAlphabet{\mathsfit}{\encodingdefault}{\sfdefault}{m}{sl}
\SetMathAlphabet{\mathsfit}{bold}{\encodingdefault}{\sfdefault}{bx}{n}

\usepackage{hyperref}
\usepackage{url}
\usepackage{amsmath, amssymb, amsfonts}
\usepackage{amsthm}
\theoremstyle{plain}
\newtheorem{proposition}{Proposition}
\newtheorem{theorem}{Theorem}

\theoremstyle{definition}

\usepackage{booktabs, multirow, array}
\usepackage{graphicx}
\usepackage{microtype}
\usepackage{caption}
\usepackage{subcaption}
\usepackage[export]{adjustbox}
\usepackage{placeins}
\usepackage{float}
\usepackage{enumitem}
\usepackage{wrapfig}
\usepackage{cleveref}
\crefname{equation}{Eq.}{Eqs.}
\Crefname{equation}{Eq.}{Eqs.}
\graphicspath{{figures/}}


\newcommand{\posdiff}[1]{\textcolor{ForestGreen}{\scriptsize \textbf{+#1}}}
\newcommand{\negdiff}[1]{\textcolor{BrickRed}{\scriptsize \textbf{-#1}}}
\newcommand{\neudiff}[1]{\textcolor{gray}{\scriptsize (#1)}}

\hypersetup{
    pdftitle={Let It Be Simple: One-Step Action Generation for Vision-Language-Action Models},
    pdfauthor={Yitong Chen, Shiduo Zhang, Jingjing Gong, Xipeng Qiu},
    pdfsubject={Vision-Language-Action Models, Flow Matching, Robot Learning},
}
\title{Let It Be Simple: One-Step Action Generation for Vision-Language-Action Models}

\author{%
Yitong Chen\textsuperscript{1,2}\quad
Shiduo Zhang\textsuperscript{2,3}\quad
Jingjing Gong\textsuperscript{2,3}\quad
Xipeng Qiu\textsuperscript{2,3}\\[0.3em]%
\mdseries\small
\textsuperscript{1}University of Science and Technology of China\quad
\textsuperscript{2}Shanghai Innovation Institute\quad
\textsuperscript{3}Fudan University\\[0.15em]%
\texttt{cyt050719@mail.ustc.edu.cn}\quad\texttt{xpqiu@fudan.edu.cn}%
}

\iclrfinalcopy
\begin{document}

\maketitle
\lhead{}  

\begin{abstract}
One-step generation from standard diffusion or Flow-Matching training is rarely effective for text-to-image or video generation without distillation or specialized one-step objectives. We find that VLA action generation behaves strikingly differently. Biasing only the training timestep distribution toward the noise endpoint yields strong one-step policies across LIBERO, LIBERO-Plus, LIBERO-Pro, and real-world robot tasks. Moreover, for open-source VLA models with different architectures, simply replacing iterative decoding with a single forward pass often performs comparably to the original multi-step decoding on benchmarks such as LIBERO, CALVIN, and SimplerEnv.

We attribute this difference to the condition-target structure of VLA generation, which is closer to image-to-text than text-to-image: rich observations strongly constrain a comparatively compact target. We formalize this intuition through the irreducible velocity loss $R_v(t,c)$ of standard Flow Matching, which becomes much smaller toward the noise endpoint for VLA, whereas image-generation models usually exhibit a U-shaped profile.

The geometry induced by the learned generative map shows the same contrast: the Jacobian of this map with respect to the initial noise has a substantially smaller largest singular value for action generation than for image generation. Together, these results suggest that VLA’s condition-target structure makes one-step generation unusually accessible to standard Flow Matching.
\end{abstract}
\section{Introduction}
\label{sec:intro}

Diffusion and flow models typically trade inference latency for sample quality via iterative numerical integration~\citep{ho2020ddpm,albergo2023stochastic,lipman2023flow}.
In text-to-image and video generation, few-step or single-step sampling remains notoriously challenging because a sparse text prompt leaves a broad, high-dimensional, and multimodal target distribution~\citep{song2023consistency,pham2026hidden}.
Consequently, effective fast generators in vision have historically necessitated specialized machinery, such as consistency training, adversarial distillation, flow-map, or mean-flow formulations~\citep{song2023consistency,yin2023dmd,frans2025shortcut,geng2025meanflow,boffi2025flowmap,lu2026onesteplatentfreeimagegeneration}.
In contrast, we discover that continuous Vision-Language-Action (VLA) flow policies~\citep{physical_intelligence_pi0_2024,physical_intelligence_pi05_2025,luo2026simvla,zheng2025xvlasoftpromptedtransformerscalable} trained under vanilla conditional flow matching natively achieve remarkable one-step performance.
Across diverse architectures and simulation and hardware benchmarks~\citep{liu2023libero,fei2025liberoplus,zhou2026liberopro,li24simpler,mees2022calvin}, a single forward pass matches or even outperforms multi-step ODE decoding, with high-noise training schedules further amplifying this capability without altering the objective.

We attribute this fundamental divergence to the underlying \emph{condition-target structure}, which governs the complexity of the residual generative problem after conditioning.
In VLA action generation, the policy receives rich multimodal streams—multi-view visual observations, natural language instructions, and proprioceptive state vectors~\citep{physical_intelligence_pi0_2024,chi2023diffusion,octo_2024}—to generate a short action trajectory.
From this perspective, VLA decoding aligns with image-to-text transcription rather than open-ended visual synthesis: extensive conditioning tightly constrains the admissible trajectory manifold to a concentrated conditional posterior.
Diagnostic ablations on LIBERO confirm this distinction: expanding the action horizon increases target complexity, which multi-step solvers readily decode; and weakening the conditioning signals disrupts the underlying condition structure, degrading closed-loop performance in ways that multi-step sampling cannot rescue.

To formalize this intuition, we analyze the irreducible velocity risk of conditional flow matching~\citep{lipman2023flow,albergo2023stochastic,liu2022flow}. Across diverse architectures and benchmarks, VLA policies consistently exhibit a monotonic loss profile that drops toward the noise boundary, unlike the U-shaped barrier in visual synthesis~\citep{li2025denoise,lu2026onesteplatentfreeimagegeneration}. A local geometric lens corroborates this contrast: while visual flows are expansive, VLA transport is predominantly contractive. Yet this contraction does not collapse the policy into deterministic regression: the learned flow retains sparse, noise-dependent spectral fractures---a structure that pure-noise endpoint training fails to recover.

We summarize our primary contributions as follows:
\begin{itemize}\setlength{\itemsep}{1.5pt}\setlength{\parskip}{0pt}\setlength{\parsep}{0pt}
    \item \textbf{Cross-Benchmark Single-Step Parity}: Across standard LIBERO~\citep{liu2023libero}, LIBERO-Plus~\citep{fei2025liberoplus}, LIBERO-Pro~\citep{zhou2026liberopro}, CALVIN~\citep{mees2022calvin}, SimplerEnv~\citep{li24simpler}, and real-world DualYam experiments~\citep{posttraining_robotics_2026}, standard flow-matching policies match multi-step numerical integration in a single step without distillation. Biasing the training timestep schedule toward high-noise states further eliminates residual discretization gaps without altering the objective.
    
    \item \textbf{Irreducible Velocity Risk and Gaussian Transport Concentration}: We formalize the condition-target dichotomy via the irreducible velocity risk $R_v(t,c)$, proving the endpoint identity $R_v(0,c)=\operatorname{Tr}\operatorname{Cov}(x_1\mid c)$. Through the pullback metric of the generative map, we establish via the Gaussian Poincar\'e inequality and concentration bounds that transport contractivity enforces exponential sub-Gaussian concentration of the target around its conditional mean. Empirically, real VLA policies exhibit a monotonic risk profile that vanishes toward the noise endpoint, contrasting sharply with the severe U-shaped barrier of visual synthesis.
    
    \item \textbf{Transport Geometry and Physical Phase Boundaries}: Evaluating the noise-to-action Jacobian $J = \nabla_{x_0} \phi_1^\theta$ demonstrates that while image flows operate in an expansive regime, VLA policies exhibit a predominantly contractive spectrum ($\sigma_1(J) \le 1$). Crucially, the sparse spectral fractures where contractivity breaks down demarcate discrete physical phase boundaries, causally steering action switches and reliably predicting gripper actuation timing.
\end{itemize}
\section{Related Work}
\label{sec:related}

\textbf{Fast sampling in flow and diffusion: from vision to action policies.}
Sampling in diffusion and continuous flow models natively relies on iterative numerical integration~\citep{ho2020ddpm,song2021score,lipman2023flow,liu2022flow}.
In visual synthesis, single-step sampling is notoriously challenging due to the under-constrained multimodal targets, prompting specialized acceleration methods such as distillation~\citep{salimans2022progressivedistillationfastsampling,yin2023dmd}, consistency models~\citep{song2023consistency}, flow maps~\citep{boffi2025flowmap,frans2025shortcut}, and mean-flow or drift formulations~\citep{geng2025meanflow,deng2026generativemodelingdrifting}.
While recent robotics efforts have similarly turned to consistency training or self-distillation to accelerate continuous VLA policies~\citep{song2025ceed,luan2026snapflow}, we find that standard flow matching natively achieves single-step parity across architectures and benchmarks—a capability further boosted by high-noise training schedules~\citep{esser2024flow,li2025denoise} without auxiliary machinery.

\textbf{Condition-target structure across visual and robotic generation.}
The tractability of conditional generation is governed not merely by raw information quantity, but fundamentally by the \emph{representational structure} of the condition relative to the target manifold.
In visual synthesis, conditioning has progressively evolved from discrete class tokens in DiT and SiT~\citep{peebles2023scalablediffusionmodelstransformers,ma2024sitexploringflowdiffusionbased} and text embeddings (CLIP, T5)~\citep{radford2021clip,raffel2020t5} to structured multimodal representations~\citep{wang2025hunyuanvideo,bai2025qwen3vltechnicalreport,chen2026scalingpropertiestextconditioning}, alongside compact target latent spaces~\citep{rombach2022ldm,dieleman2025latents,yao2025vavae,tian2024var,yu2023magvit,yu2024magvitv2}.
Unlike visual synthesis, VLA policies~\citep{physical_intelligence_pi0_2024,physical_intelligence_pi05_2025,luo2026simvla,zheng2025xvlasoftpromptedtransformerscalable,chi2023diffusion,octo_2024} leverage rich multimodal context to predict an intrinsically compact action trajectory.
Consequently, grounded by structured multimodal conditions, this compact action manifold allows standard flow matching to decode effectively in a single step.

\textbf{Irreducible velocity risk and Jacobian spectral diagnostics.}
Time-dependent vector field optimization in flow matching is fundamentally governed by the Bayes conditional variance along the probability path~\citep{lipman2023flow,albergo2023stochastic,liu2022flow}.
In diffusion and flow matching, an irreducible velocity risk barrier emerges when the conditioning signal fails to resolve uncertainty over an expansive target space, as observed in under-constrained pixel diffusion~\citep{li2025denoise,lu2026onesteplatentfreeimagegeneration}.
In contrast, when the condition structure is rich relative to the target—or when the target manifold is intrinsically compact, as in VLA policies—this ambiguity collapses, and the resulting velocity loss is empirically monotonic.
Furthermore, while sensitivity and spectrum analysis of generative maps $J = \nabla_{x_0} \phi_1^\theta$ have been studied for controllable editing~\citep{dflow-pmlr-v235-ben-hamu24a,chen2024low-dim-control-diffusion} and out-of-distribution detection~\citep{2026eigenscoreood}, we extend this geometric lens to continuous action policies.
Remarkably, this unifying perspective reveals that just as Jacobian singular values capture discrete semantic transitions in visual generation, their spectral ridges in VLA directly demarcate physical phase boundaries—such as contact transitions and gripper actuations—while confirming the contractive nature of action flows.

\section{From Velocity Risk to Gaussian Transport Concentration}
\label{sec:risk}

\subsection{Flow Matching Preliminaries}

Let $\mathbb{R}^d$ denote the data space. Flow matching~\citep{lipman2023flow,liu2022flow,albergo2023stochastic} models continuous generative trajectories by regressing a time-dependent vector field $v_\theta(x, t, c)$ to generate a probability path $p_t$ bridging a prior $p_0 = \mathcal{N}(0, I)$ at $t=0$ to the data distribution $p_1$ at $t=1$ via the ODE $\frac{d}{dt}\phi_t(x) = v_t(\phi_t(x))$.
Because regressing the marginal vector field $\mathbb{E}_{t, p_t}[\|v_\theta - u_t\|^2]$ is generally intractable, Conditional Flow Matching (CFM) trains the model on conditional probability paths $p_t(x \mid x_1)$ generated by per-sample conditional vector fields $u_t(x \mid x_1)$, which yields identical gradients with respect to $\theta$.

In practice, we adopt the standard linear interpolation $x_t = t x_1 + (1-t) x_0$ with $x_0 \sim \mathcal{N}(0, I)$, which yields a constant conditional velocity $u_t(x_t \mid x_1) = x_1 - x_0$. The policy is trained by optimizing $
    \mathcal{L}_{\text{CFM}}(\theta) = \mathbb{E}_{t \sim \mathcal{U}[0,1], \, x_0 \sim \mathcal{N}(0,I), \, (c, x_1) \sim q} \left\| v_\theta(x_t, t, c) - (x_1 - x_0) \right\|^2.$

\subsection{Irreducible Velocity Risk}

Under the linear interpolation path, the flow matching loss is lower-bounded by the Bayes conditional variance: $R_v(t,c) \triangleq \mathbb{E} \operatorname{Tr} \operatorname{Var}(x_1 - x_0 \mid x_t, c) = \frac{1}{(1-t)^2} \mathbb{E} \operatorname{Tr} \operatorname{Cov}(x_1 \mid x_t, c)$.
At $t=0$, this simplifies to the target dispersion $R_v(0,c) = \operatorname{Tr} \operatorname{Cov}(x_1 \mid c)$. Since the endpoint field targets the conditional mean, $v^*(x_0, 0, c) = \mathbb{E}[x_1 \mid c] - x_0$, a single Euler step predicts $\mathbb{E}[x_1 \mid c]$ with expected distortion $R_v(0,c)$. Single-step viability thus hinges on whether conditioning concentrates the posterior measure sufficiently to render this mean a faithful trajectory proxy.

\begin{figure}[!htbp]
    \centering
    \includegraphics[width=\linewidth]{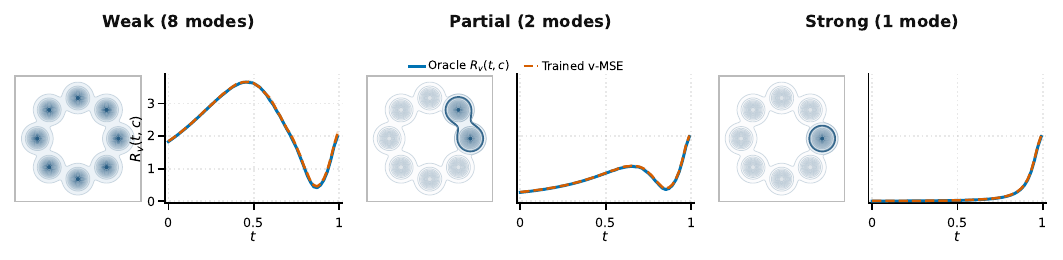}
    \caption{Each column shows the target geometry (left) and the analytical/trained risk profiles (right), from weak to strong conditioning.}
    \label{fig:cond_target_profile}
\end{figure}
For targets distributed as an $M$-component Gaussian mixture $x_1 \sim \sum_k \pi_k \mathcal{N}(\mu_k, \sigma^2 I)$, the law of total variance yields a closed-form posterior covariance along the linear-Gaussian path:
\[
    \operatorname{Cov}(x_1 \mid x_t, c) = \frac{\sigma^2(1-t)^2}{(1-t)^2 + t^2\sigma^2} I + \sum_{k \in \mathcal{S}_c} w_k(x_t) (\mu_{k,t} - \bar{\mu}_t)(\mu_{k,t} - \bar{\mu}_t)^\top,
\]
where $\mathcal{S}_c$ denotes the subset of modes consistent with condition $c$, $w_k(x_t) = p(k \mid x_t, c)$ is the posterior component probability, and $\mu_{k,t} = \mathbb{E}[x_1 \mid x_t, k]$.

Figure~\ref{fig:cond_target_profile} visualizes this contraction: across all three regimes, an identical MLP tracks the analytical Bayes-risk profile, showing that stronger conditioning directly lowers endpoint velocity risk.

\subsection{Pullback Geometry and Gaussian Concentration}
\label{sec:theory_jacobian}

While the scalar trace $R_v(t, c)$ tracks aggregate posterior variance, it inherently overlooks the anisotropic geometry of the conditioned target.
To resolve this, we leverage the differential properties of the transport map $\phi_1(\cdot; c)$ as an intrinsic probe of target manifold regularity.

The Euclidean metric on the target action space induces a Riemannian pullback metric $g(x_0; c) \triangleq J(x_0; c)^\top J(x_0; c)$ on the prior noise domain via the generative map $\phi_1$, where $J(x_0; c) \triangleq \nabla_{x_0} \phi_1(x_0; c)$.
The operator norm $\sigma_1(J) = \sqrt{\lambda_{\max}(g)}$ quantifies the maximal local expansion under push-forward.
Applying the Gaussian Poincar\'e inequality $\operatorname{Var}_{\gamma}[f] \le \mathbb{E}_{\gamma}[\|\nabla f\|^2]$ coordinate-wise under the prior $\gamma = \mathcal{N}(0, I_d)$ directly bounds the global target dispersion via
\[\operatorname{Tr}\!\operatorname{Cov}(x_1 \mid c) \le \sum_{i=1}^d \mathbb{E}_{x_0}[\|\nabla_{x_0} \phi_1^{(i)}(x_0; c)\|^2] = \mathbb{E}_{x_0}[\|J(x_0; c)\|_F^2] \le d\,\mathbb{E}_{x_0}[\sigma_1(J(x_0; c))^2],\]
so an on-average contractive map ($\mathbb{E}[\sigma_1^2] \le 1$) directly guarantees a low-variance target by compressing the aggregate trace covariance relative to the prior.

While this trace bound controls average dispersion, a uniform spectral bound $L(c) \triangleq \operatorname*{ess\,sup}_{z}\|J(z;c)\|_2$ further upgrades this control to exponential concentration. Via Gaussian concentration (Herbst argument; Appendix~\ref{app:proof_subgaussian}), we obtain
\begin{equation}
\label{eq:subgaussian_norm}
\mathbb{P}\left( \|x_1 - \mathbb{E}[x_1 \mid c]\|_2 \ge \sqrt{\mathbb{E}_z\|J(z;c)\|_F^2} + \epsilon \middle| c \right) \le \exp\left( -\frac{\epsilon^2}{2L(c)^2} \right), \quad \forall \epsilon > 0.
\end{equation}
Here $\sqrt{\mathbb{E}_z\|J(z;c)\|_F^2}\le L(c)\sqrt d$ controls the characteristic radius, while $L(c)$ controls the exponential tail; when $L(c)\le1$, all centered one-dimensional projections are sub-Gaussian with variance proxy at most $1$.

\section{Continuous Vision-Language-Action Flow Policies}
\label{sec:vla}

\subsection{Controlled Limit Case: Deterministic Target with Complex Condition}
\label{sec:toy_limit_case}

To verify that low-variance target manifolds natively admit single-step flow matching before confronting robotics dataset, we first probe a synthetic limit case: \textbf{the target posterior collapses to a Dirac delta, while the condition requires parsing high-dimensional visual scenes.}
We instantiate this by mapping an image of a multi-digit MNIST grid ($c$) to its deterministic token sequence ($x_1$) using a standard encoder--decoder flow model (Fig.~\ref{fig:mnist_all}(a); details in Appendix~\ref{app:mnist}).

Consistent with our risk analysis ($R_v(0, c) \to 0$), empirical velocity MSE minimizes at the noise boundary $t=0$ (Fig.~\ref{fig:mnist_all}b). Biasing training time toward noise ($t \sim \mathrm{Beta}(0.5, 1.0)$) lifts one-step decoding to 10-step Euler parity—indeed, pure-noise training succeeds on MNIST (Fig.~\ref{fig:mnist_all}c) but fails entirely on CIFAR-10 (Fig.~\ref{fig:mnist_all}d), confirming that target posterior concentration natively enables one-step flow matching.

This limit case also dictates our architecture. Encoder--decoder scaling ablations show that a weaker vision encoder degrades one-step performance, while expanding decoder capacity fails to compensate (Appendix Fig.~\ref{fig:mnist_ablations}). Thus, the bottleneck lies strictly in condition representation—motivating a powerful vision-language backbone paired with a lightweight flow action head.

\begin{figure}[!htbp]
    \centering
    \begin{subfigure}[t]{0.18\linewidth}
        \centering
        \includegraphics[width=\linewidth]{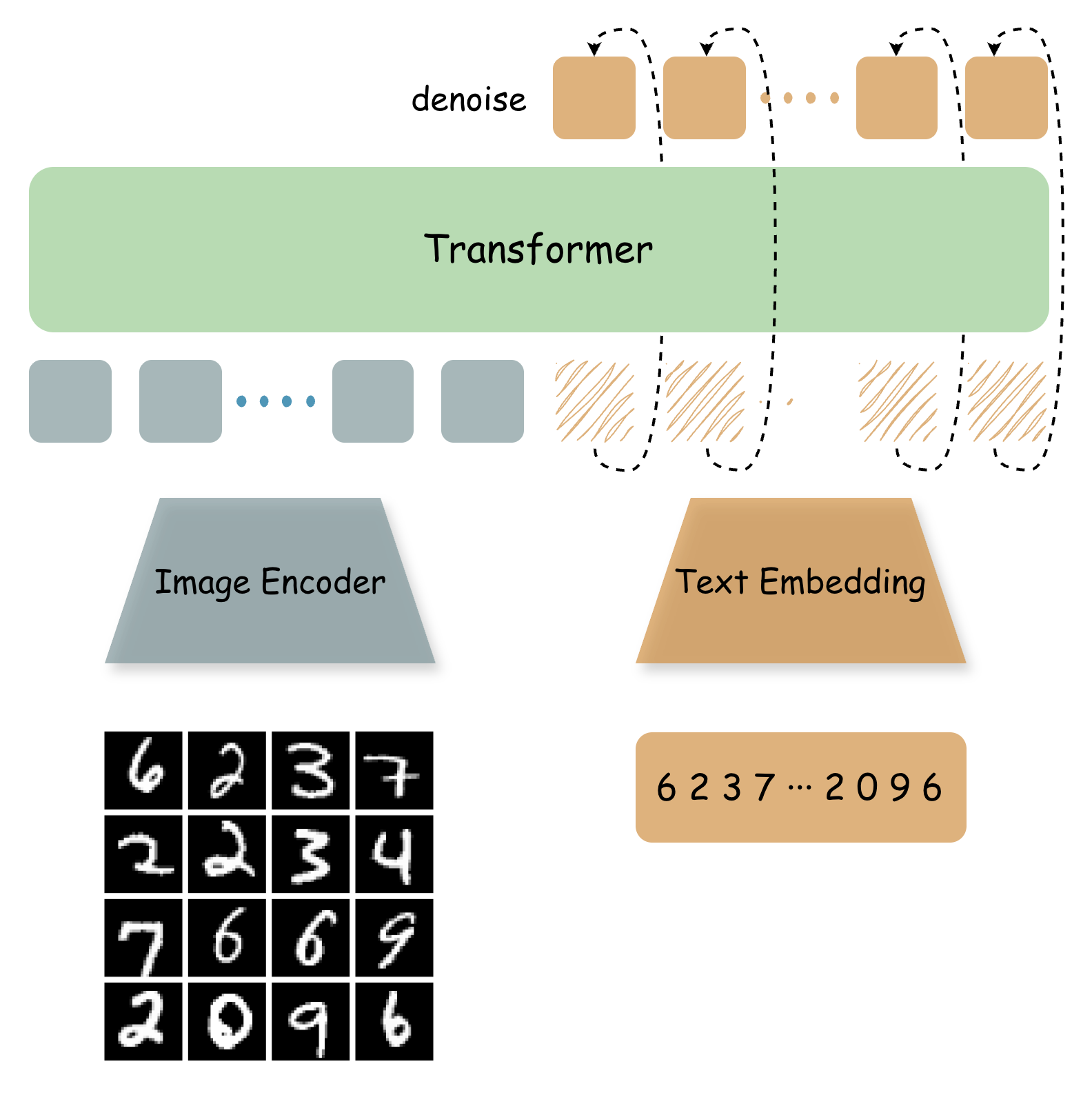}
        \caption{Task \& model.}
        \label{fig:mnist_grid}
    \end{subfigure}\hfill
    \begin{subfigure}[t]{0.27\linewidth}
        \centering
        \includegraphics[width=\linewidth]{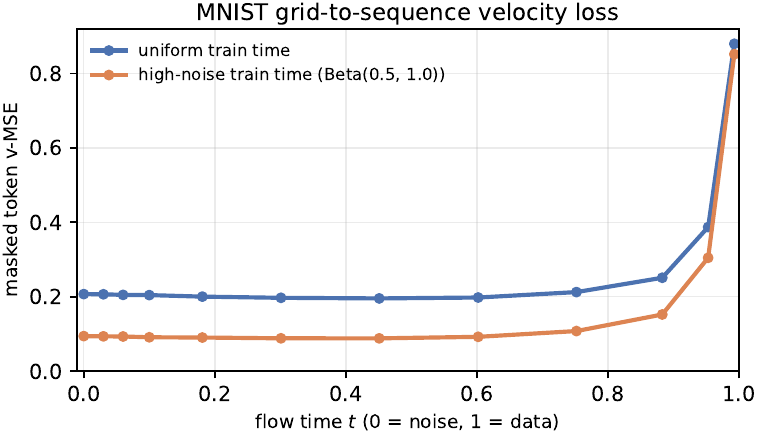}
        \caption{v-loss profile.}
        \label{fig:mnist_vloss}
    \end{subfigure}\hfill
    \begin{subfigure}[t]{0.26\linewidth}
        \centering
        \includegraphics[width=\linewidth]{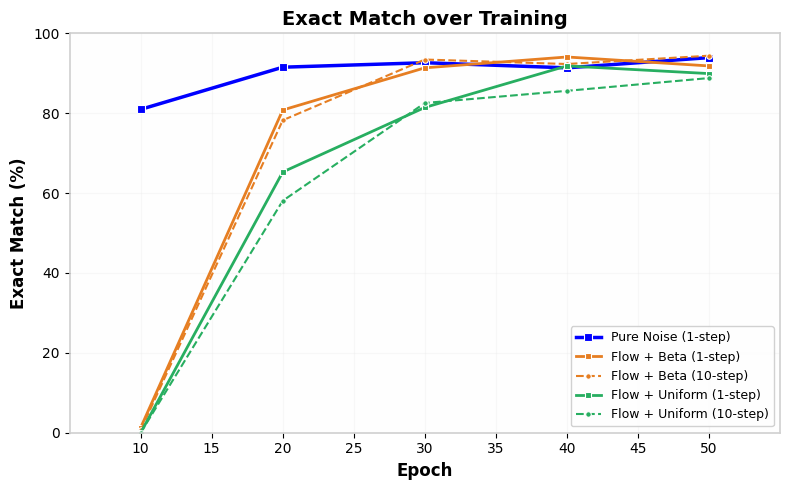}
        \caption{One-step exact match.}
        \label{fig:mnist_schedule}
    \end{subfigure}\hfill
    \begin{subfigure}[t]{0.26\linewidth}
        \centering
        \includegraphics[width=\linewidth]{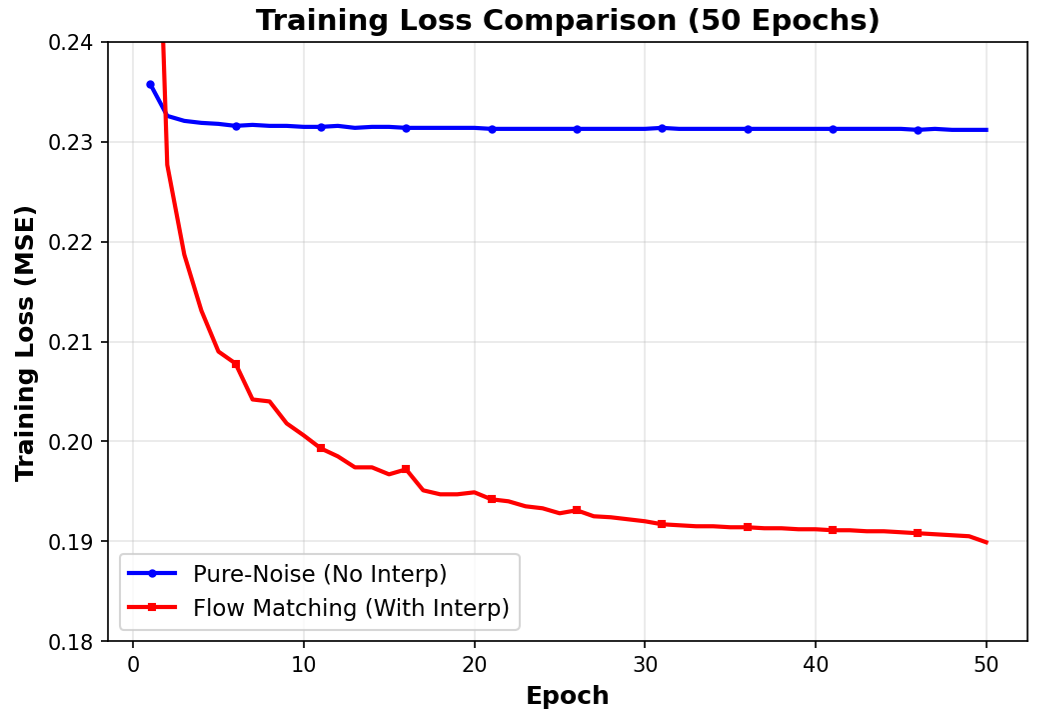}
        \caption{CIFAR-10 contrast.}
        \label{fig:mnist_cifar10}
    \end{subfigure}
    \caption{Diagnostics on image-to-text MNIST (a--c) and class-to-image CIFAR-10 (d). (a) Input grid. (b) Empirical velocity MSE over flow time $t$. (c) Exact-match accuracy across training time distributions (uniform, high-noise Beta, and pure noise). (d) Pure-noise training loss contrast.}
    \label{fig:mnist_all}
\end{figure}

\subsection{Embodied Policy Architecture and Action Flow Parameterization}
\label{sec:vla_arch}

Robot manipulation operates in this same favorable regime: rich multimodal context (camera streams, language commands, and proprioceptive state vectors) tightly bounds an intrinsically compact target---a short action chunk $x_1 \in \mathbb{R}^{H \times d_a}$ over horizon $H$.

\begin{wrapfigure}{r}{0.46\linewidth}
    \centering
    \includegraphics[width=\linewidth]{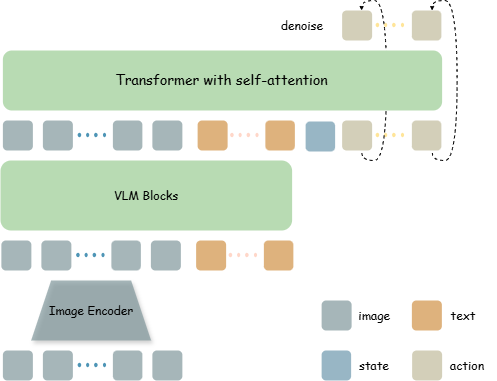}
    \caption{Overview of the flow-based VLA architecture: multimodal condition tokens modulate a lightweight action head that regresses the action trajectory velocity field.}
    \label{fig:vla-model-arch}
\end{wrapfigure}

\textbf{Architecture Guided by Condition Dominance.}
Grounded by our finding that generative performance is bottlenecked by condition representation rather than target-side capacity, our architecture decouples high-capacity multimodal perception from lightweight action decoding (Fig.~\ref{fig:vla-model-arch}). Built upon OpenPI~\citep{physical_intelligence_pi0_2024}, a vision-language backbone processes multi-camera observations via SigLIP~\citep{zhai2023siglip} and fuses them with instruction tokens through PaliGemma~\citep{beyer2024paligemma} to produce contextual prefix tokens. A lightweight bidirectional Transformer head~\citep{luo2026simvla} takes these prefix tokens along with the robot proprioceptive state $s_t$, flow time $t$, and noised action candidate $x_t \in \mathbb{R}^{H \times d_a}$ to predict velocity $v_\theta(x_t, t, c)$.

We evaluate across two model capacities while fixing the action head at $\sim$32M parameters: a \emph{Tiny} variant ($\sim$109M active parameters) for extensive diagnostic sweeps, and a \emph{Full} variant ($\sim$888M active parameters) to examine scaling behavior (architectural breakdown in Appendix~\ref{app:robot_setup}).

\section{Robot Manipulation Experiments}
\label{sec:robot}

\subsection{High-Noise Training Unlocks One-Step VLA Policies}
\label{sec:libero_schedule}
\enlargethispage{2\baselineskip}

\begin{wraptable}{r}{0.62\linewidth}
\centering
\captionsetup{font=footnotesize}
\caption{\textbf{Time-schedule controls on LIBERO} (success rates \%). Differences ($\pm$) are relative to Uniform 1-step; best 1-step results are \textbf{bolded}.}
\label{tab:time_ablation}
\footnotesize
\setlength{\tabcolsep}{4pt}
\renewcommand{\arraystretch}{0.9}
\begin{tabular}{llcccc}
\toprule
\textbf{Schedule} & \textbf{Steps} & \textbf{Spatial} & \textbf{Object} & \textbf{Goal} & \textbf{Long} \\
\midrule
Uniform & 10 & 96.6 & 96.2 & 93.2 & 80.8 \\
Uniform & 1  & 88.8 & 92.8 & 90.2 & 70.2 \\
\midrule
$\alpha=3$ & 1  & 95.8 \posdiff{7.0} & 97.0 \posdiff{4.2} & 94.4 \posdiff{4.2} & 78.0 \posdiff{7.8} \\
$\alpha=3$ & 10 & 91.6 & 96.2 & 82.2 & 52.4 \\
\midrule
$\alpha=4$ & 1  & \textbf{96.4} \posdiff{7.6} & \textbf{99.6} \posdiff{6.8} & \textbf{96.8} \posdiff{6.6} & \textbf{85.2} \posdiff{15.0} \\
$\alpha=4$ & 10 & 93.4 & 96.8 & 91.4 & 63.4 \\
\midrule
$\alpha=8$ & 1  & 95.6 \posdiff{6.8} & 99.0 \posdiff{6.2} & 94.0 \posdiff{3.8} & 78.0 \posdiff{7.8} \\
$\alpha=8$ & 10 & 78.4 & 40.0 & 56.6 & 22.6 \\
\midrule
Pure-noise & 1  & 96.0 \posdiff{7.2} & 97.8 \posdiff{5.0} & 96.2 \posdiff{6.0} & 72.8 \posdiff{2.6} \\
\bottomrule
\end{tabular}
\end{wraptable}

Evaluating noise-shifted training ($t = u / [1 + (\alpha-1)(1-u)]$, $\alpha \ge 1$) across LIBERO suites~\citep{liu2023libero} (Table~\ref{tab:time_ablation}) reveals that \textbf{biasing training toward noise ($\alpha=4$) eliminates the one-step gap}, consistently outperforming the 10-step uniform baseline across all four suites. Naturally, over-weighting the noise endpoint sacrifices intermediate path accuracy, degrading 10-step integration under extreme bias ($\alpha=8$). Indeed, whereas image flow matching hinges on noise-data interpolation to optimize complex distributions, the success of pure-noise training highlights that strong conditioning places embodied control in a fundamentally different geometric regime.

\begin{wraptable}{r}{0.56\linewidth}
\centering
\vspace{-\intextsep}
\captionsetup{font=footnotesize}
\caption{\textbf{Full-encoder scaling performance} ($\alpha=3$, success rates \%). One-step decoding matches 10-step Euler integration across all suites. Differences ($\pm$) are relative to 10-step.}
\label{tab:full_encoder}
\footnotesize
\setlength{\tabcolsep}{4pt}
\renewcommand{\arraystretch}{0.9}
\begin{tabular}{lcccc}
\toprule
\textbf{Steps} & \textbf{Spatial} & \textbf{Object} & \textbf{Goal} & \textbf{Long} \\
\midrule
1-step  & 98.4 \neudiff{-0.4} & 100.0 \posdiff{1.0} & 97.0 \neudiff{-1.0} & 95.6 \posdiff{0.6} \\
10-step & 98.8 &  99.0 & 98.0 & 95.0 \\
\bottomrule
\end{tabular}
\end{wraptable}

\paragraph{Encoder Scaling and the Necessity of Prior Geometry.}
Scaling the perceptual encoder to the Full variant pushes one-step decoding to full parity with 10-step Euler integration across all suites (Table~\ref{tab:full_encoder}). 
Yet, we also try the direct regression: training a Tiny variant that completely replaces prior noise with learnable input embeddings degrades substantially on complex suites (Appendix Table~\ref{tab:learned_start}).
This gap indicates that the target distribution retains nontrivial structure, motivating our geometric analysis in Section~\ref{sec:theory_jacobian}.

\begin{wraptable}{r}{0.60\linewidth}
\centering
\vspace{-\intextsep}
\captionsetup{font=footnotesize}
\caption{\textbf{Paired decoding of official $\pi$ checkpoints on LIBERO} (success rates \%, 500 episodes/suite). Differences ($\pm$) are relative to each checkpoint's 10-step baseline.}
\label{tab:paired_decoding}
\footnotesize
\setlength{\tabcolsep}{4pt}
\renewcommand{\arraystretch}{0.9}
\begin{tabular}{llcccc}
\toprule
\textbf{Model} & \textbf{Steps} & \textbf{Spatial} & \textbf{Object} & \textbf{Goal} & \textbf{Long} \\
\midrule
$\pi_{0.5}$ & 10 & 99.0 & 99.4 & 98.0 & 94.2 \\
$\pi_{0.5}$ & 1 & 99.4 \posdiff{0.4} & 99.4 \neudiff{0.0} & 98.8 \posdiff{0.8} & 95.6 \posdiff{1.4} \\
\midrule
$\pi_0$ & 10 & 95.4 & 97.6 & 96.0 & 85.6 \\
$\pi_0$ & 1 & 95.4 \neudiff{0.0} & 97.6 \neudiff{0.0} & 96.0 \neudiff{0.0} & 82.2 \negdiff{3.4} \\
\bottomrule
\end{tabular}
\end{wraptable}

\paragraph{Generalization Across Diverse Policies.}
This behavior also holds across diverse public checkpoints~\citep{zheng2025xvlasoftpromptedtransformerscalable,ye2026starvla,physical_intelligence_pi0_2024,physical_intelligence_pi05_2025}, where one-step decoding consistently matches multi-step integration across LIBERO (Table~\ref{tab:paired_decoding}), SimplerEnv WidowX~\citep{li24simpler}, and CALVIN~\citep{mees2022calvin} (Table~\ref{tab:cross_benchmark}). This suggests that near-lossless single-step execution is an inherent property of strongly conditioned action flows.
Single-step decoding also cuts per-query model latency $1.5$--$3.4\times$ across architectures, with the condition prefix dominating the remainder (Appendix~\ref{app:latency}).
\begin{table}[!htbp]
\centering
\caption{\textbf{Cross-benchmark paired decoding of third-party checkpoints.} Evaluation on SimplerEnv WidowX~\citep{li24simpler} (24 episodes/task) and CALVIN ABC$\to$D~\citep{mees2022calvin} (subgoal success \%; Avg.\ chain length / 5). Differences ($\pm$) are relative to each checkpoint's multi-step baseline.}
\label{tab:cross_benchmark}
\footnotesize
\setlength{\tabcolsep}{3.5pt}
\renewcommand{\arraystretch}{0.95}
\resizebox{\linewidth}{!}{%
\begin{tabular}{@{}l ccccc @{\hspace{8pt}} ccccc c@{}}
\toprule
& \multicolumn{5}{c}{\textbf{SimplerEnv WidowX}} & \multicolumn{6}{c}{\textbf{CALVIN ABC$\to$D}} \\
\cmidrule(lr){2-6} \cmidrule(l){7-12}
\textbf{Model} & \textbf{Spoon} & \textbf{Carrot} & \textbf{Cube} & \textbf{Eggpl.} & \textbf{Avg.} & \textbf{LH-1} & \textbf{LH-2} & \textbf{LH-3} & \textbf{LH-4} & \textbf{LH-5} & \textbf{Avg.} \\
\midrule
Qwen2.5-$\pi$ (4-step) & 79.2 & 58.3 & 37.5 & 83.3 & \textbf{64.6} & 91.4 & 83.0 & 74.6 & 67.0 & 59.3 & \textbf{3.75} \\
Qwen2.5-$\pi$ (1-step) & 91.7 \posdiff{12.5} & 58.3 \neudiff{0.0} & 33.3 \negdiff{4.2} & 70.8 \negdiff{12.5} & \textbf{63.5 \neudiff{-1.1}} & 91.9 \posdiff{0.5} & 83.7 \posdiff{0.7} & 74.7 \posdiff{0.1} & 66.5 \neudiff{-0.5} & 57.5 \negdiff{1.8} & \textbf{3.74 \neudiff{-0.01}} \\
\addlinespace[0.3em]
X-VLA (10-step) & 95.8 & 91.7 & 87.5 & 87.5 & \textbf{90.6} & 95.6 & 91.8 & 87.1 & 81.7 & 73.9 & \textbf{4.30} \\
X-VLA (1-step) & 100.0 \posdiff{4.2} & 100.0 \posdiff{8.3} & 87.5 \neudiff{0.0} & 87.5 \neudiff{0.0} & \textbf{93.8 \posdiff{3.2}} & 95.0 \neudiff{-0.6} & 91.5 \neudiff{-0.3} & 86.4 \neudiff{-0.7} & 80.6 \neudiff{-1.1} & 72.4 \negdiff{1.5} & \textbf{4.26 \neudiff{-0.04}} \\
\bottomrule
\end{tabular}}
\end{table}

\begin{wraptable}{r}{0.54\linewidth}
\centering
\vspace{-0.8em}
\captionsetup{font=footnotesize}
\caption{\textbf{Real-world bimanual task success} (success / trials, 15 trials/task; Fig.~\ref{fig:yam}). 10-step results are from the workshop official baseline.}
\label{tab:rss_real_robot}
\footnotesize
\setlength{\tabcolsep}{8pt}
\renewcommand{\arraystretch}{0.95}
\begin{tabular}{lcc}
\toprule
\textbf{Task} & \textbf{1-step} & \textbf{10-step} \\
\midrule
Insert mouse battery   & 80.0\% & 80.0\% \\
Seal bottle cap        & 46.7\% & 35.0\% \\
Tower of Hanoi         & 40.0\% & 50.0\% \\
\bottomrule
\end{tabular}
\vspace{-1.0em}
\end{wraptable}
\vspace{-1.2em}
\paragraph{Real-World Bimanual Manipulation.}
We further validate one-step decoding on physical hardware using the dual-arm Yam platform across three tasks~\citep{posttraining_robotics_2026} (Fig.~\ref{fig:yam}). For each task, we fine-tune a policy from OpenPI $\pi_{0.5}$~\citep{physical_intelligence_pi05_2025} using 10 hours of human teleoperation demonstrations. One-step decoding matches or rivals the official 10-step baseline (Table~\ref{tab:rss_real_robot}), demonstrating real-world viability.

We show this parity persists under perturbations: one-step decoding consistently matches or even outperforms multi-step integration across LIBERO-Plus~\citep{fei2025liberoplus} and LIBERO-Pro~\citep{zhou2026liberopro} (Appendix~\ref{app:libero_plus}; numerical solver audit in Appendix Fig.~\ref{fig:solver_audit}).

\subsection{Velocity Loss Landscape Across Flow Time}
\label{sec:vloss_illustration}

\begin{figure}[t]
    \centering
    \begin{subfigure}[t]{0.24\linewidth}
        \centering
        \includegraphics[width=\linewidth]{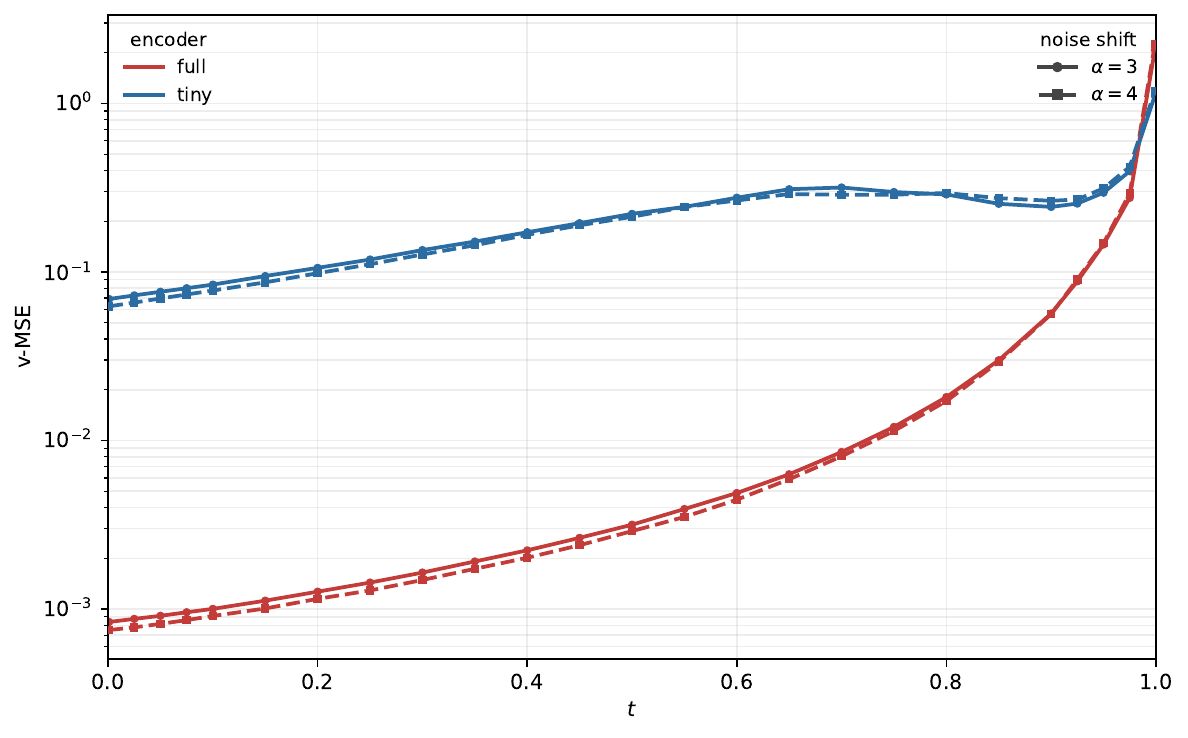}
        \caption{Encoder scale.}
        \label{fig:vloss_encoder}
    \end{subfigure}\hfill
    \begin{subfigure}[t]{0.24\linewidth}
        \centering
        \includegraphics[width=\linewidth]{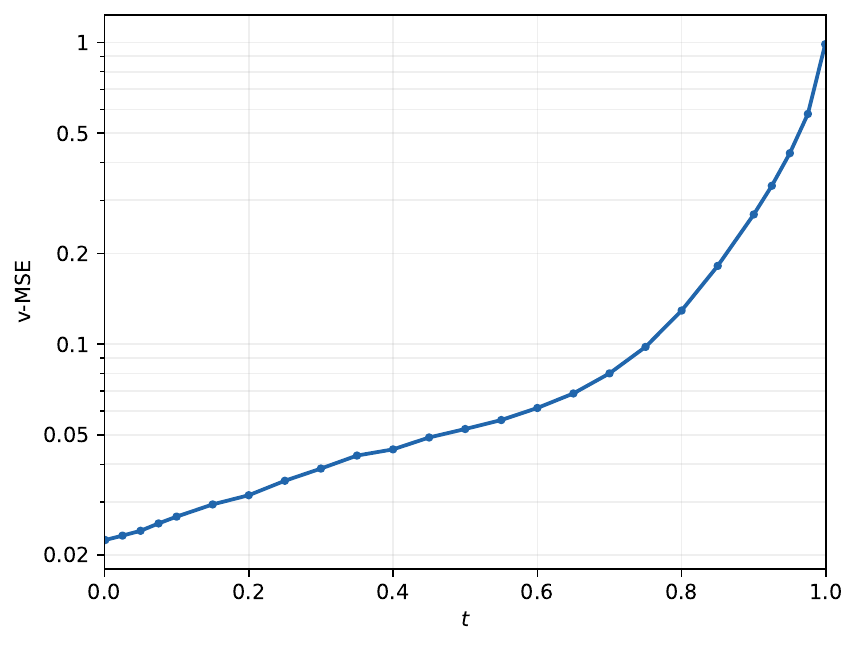}
        \caption{$\pi_{0.5}$ LIBERO.}
        \label{fig:vloss_pi05}
    \end{subfigure}\hfill
    \begin{subfigure}[t]{0.24\linewidth}
        \centering
        \includegraphics[width=\linewidth]{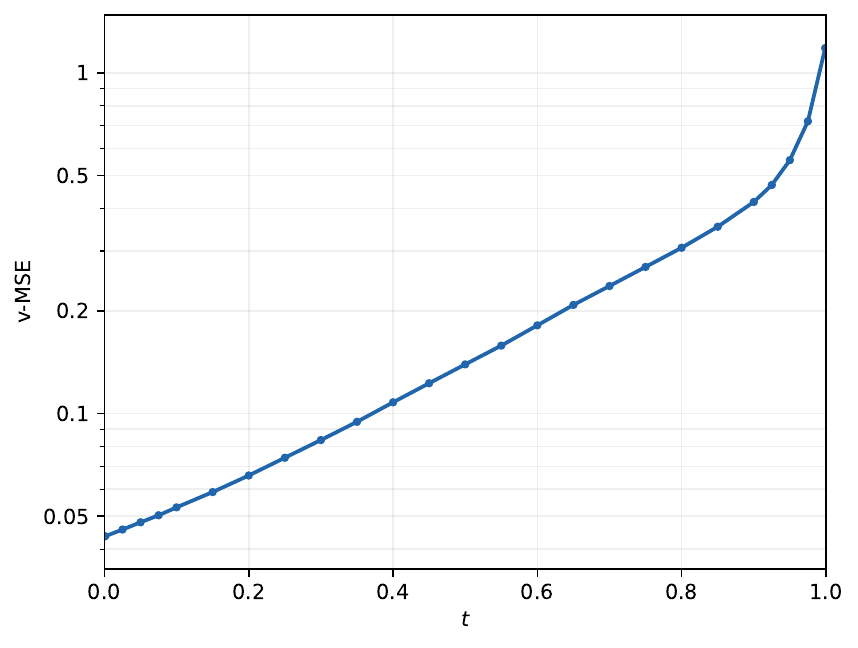}
        \caption{$\pi_0$ LIBERO-Plus.}
        \label{fig:vloss_pi0}
    \end{subfigure}\hfill
    \begin{subfigure}[t]{0.24\linewidth}
        \centering
        \includegraphics[width=\linewidth]{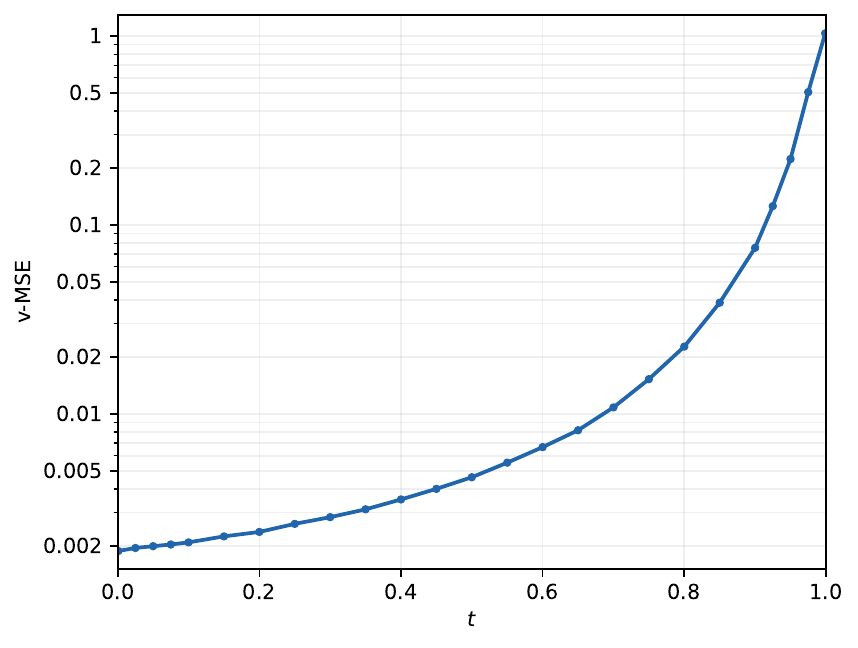}
        \caption{StarVLA LIBERO.}
        \label{fig:vloss_starvla}
    \end{subfigure}

    \vspace{0.6em}
    \begin{subfigure}[t]{0.24\linewidth}
        \centering
        \includegraphics[width=\linewidth]{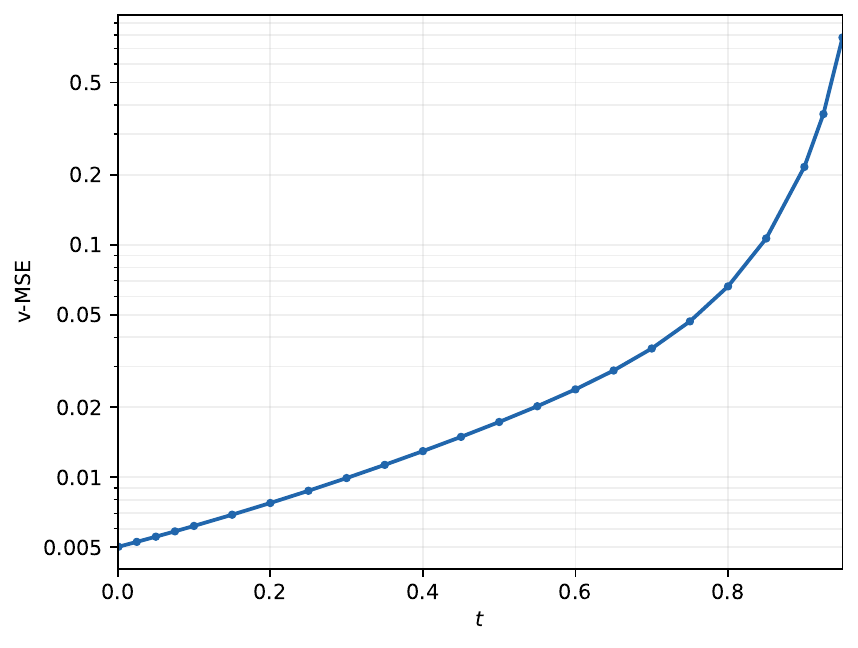}
        \caption{X-VLA CALVIN.}
        \label{fig:vloss_xvla_calvin}
    \end{subfigure}\hfill
    \begin{subfigure}[t]{0.24\linewidth}
        \centering
        \includegraphics[width=\linewidth]{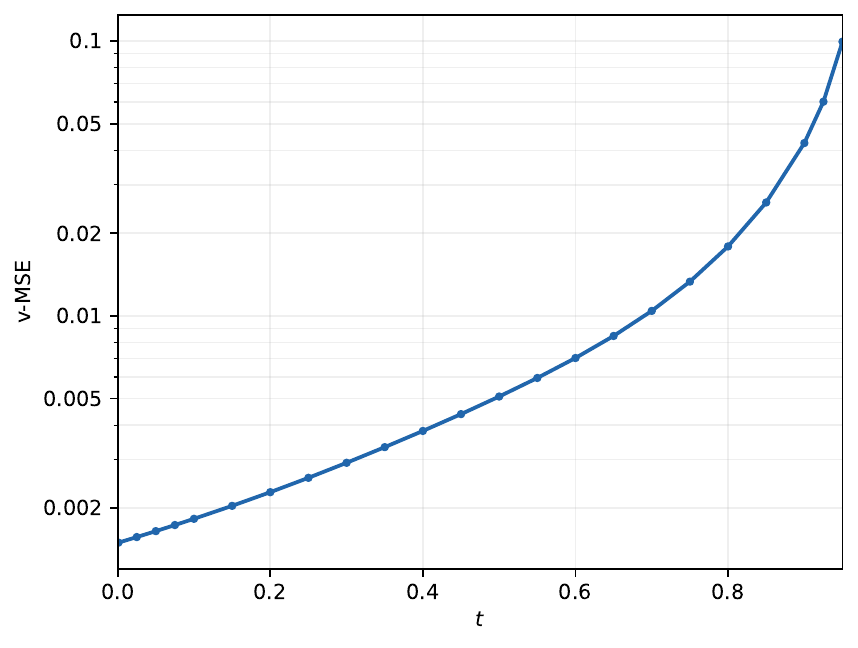}
        \caption{X-VLA WidowX.}
        \label{fig:vloss_xvla_widowx}
    \end{subfigure}\hfill
    \begin{subfigure}[t]{0.24\linewidth}
        \centering
        \includegraphics[width=\linewidth]{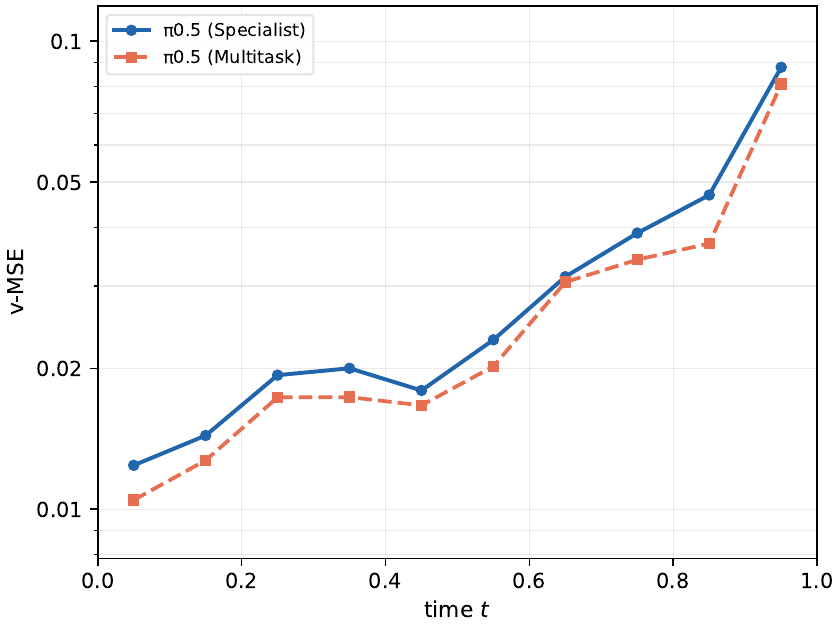}
        \caption{$\pi_{0.5}$ real robot.}
        \label{fig:vloss_rss}
    \end{subfigure}\hfill
    \begin{subfigure}[t]{0.24\linewidth}
        \centering
        \includegraphics[width=\linewidth]{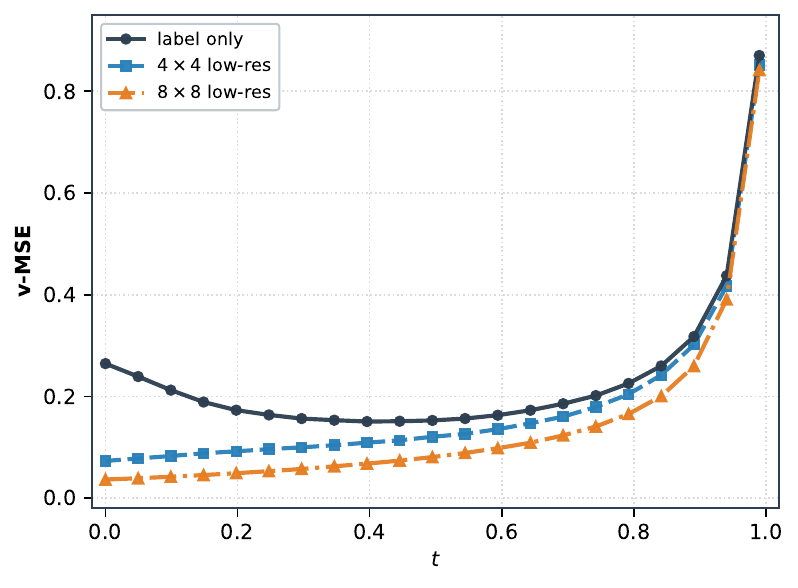}
        \caption{CIFAR-10.}
        \label{fig:vloss_cifar10}
    \end{subfigure}
    \caption{\textbf{Fixed-time velocity MSE across flow time $t \in [0, 1]$}. (a--g) Robotic policies across diverse backbones and benchmarks. (h) CIFAR-10 under varying conditioning (class label vs.\ $4\times4$ / $8\times8$ low-resolution context). Scaling the vision-language backbone uniformly lowers the error profile ((a), Tiny vs.\ Full).}
    \label{fig:vloss_profiles}
\end{figure}

Across diverse VLA architectures and benchmarks (Fig.~\ref{fig:vloss_profiles}a--g), velocity MSE increases monotonically from the noise to the clean target, matching our $R_v(t,c)$ formulation (Section~\ref{sec:risk}).
The CIFAR-10 ablation (Fig.~\ref{fig:vloss_profiles}h) explains this transition: while weak class conditioning produces a U-shaped profile, progressively adding low-resolution spatial context flattens the curve into the same monotonic profile observed in VLAs.
Furthermore, scaling from the Tiny to the Full encoder uniformly suppresses the entire loss curve (Fig.~\ref{fig:vloss_profiles}a), confirming that capacity scaling directly reduces vector field residual to boost single-step performance.

\subsection{Expanding Action Horizons and Single-Step Viability}
\label{sec:action_horizon}
\begin{figure}[!htbp]
\centering
\begin{subfigure}[t]{0.55\linewidth}
    \centering
    \includegraphics[width=\linewidth]{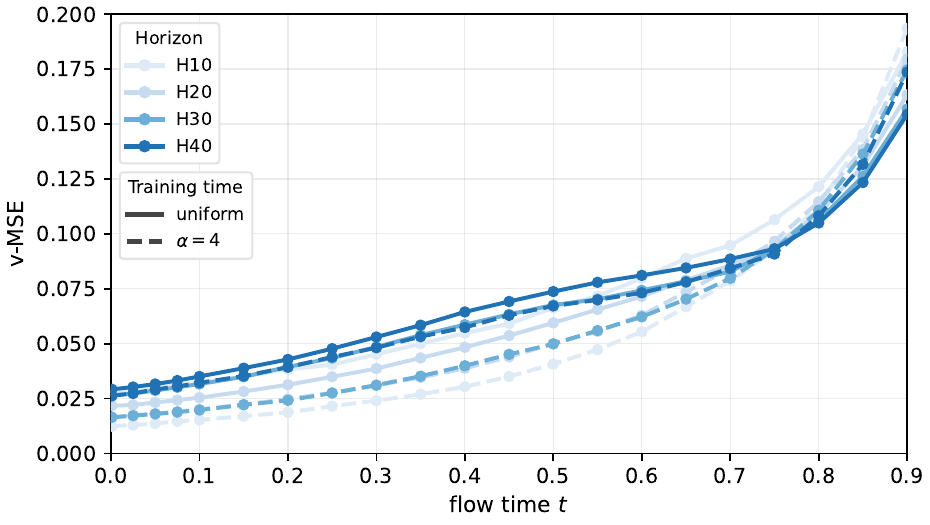}
    \phantomcaption\label{fig:horizon_vloss}
\end{subfigure}\hfill
\begin{subfigure}[t]{0.42\linewidth}
    \centering
    \includegraphics[width=\linewidth]{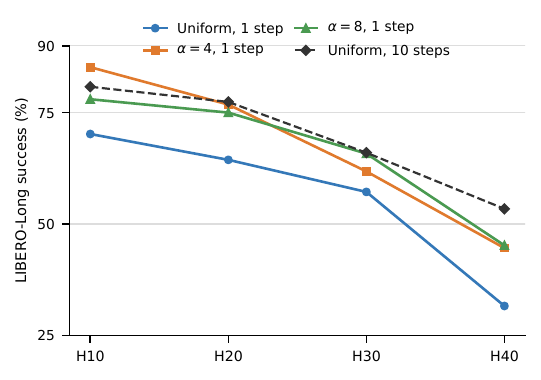}
    \phantomcaption\label{fig:horizon_trends}
\end{subfigure}
\caption{\textbf{Action horizon scaling on LIBERO-Long.} (a) Velocity MSE profiles under uniform and noise-shifted schedules across horizons $H \in \{10, 20, 30, 40\}$. (b) Single-step advantage narrows as horizon grows, with multi-step integration overtaking it at larger $H$.}
\label{fig:horizon_stress_test}
\end{figure}

We test the limits of single-step viability by extending the action horizon from $H=10$ to $H \in \{20, 30, 40\}$ on LIBERO-Long (Fig.~\ref{fig:horizon_trends}).
As the chunk length grows, single-step decoding degrades systematically: while single-step execution dominates at standard horizons ($H=10$), multi-step integration overtakes it at larger $H$, with noise shifting ($\alpha=4$) delaying this crossover.

While velocity profiles preserve the familiar early-time minimum across all $H$ (Fig.~\ref{fig:horizon_vloss}), loss magnitudes do not cleanly rank with horizon length.
Although velocity loss serves as a useful proxy for endpoint risk, it conflates optimization error with action dimensionality across horizons. 
We complement it with the local Jacobian spectrum ($\sigma_1$) to examine the flow sensitivity directly.

\section{Empirical Transport Geometry: Global Contraction and Local Phase Boundaries}
\label{sec:geometry}

\subsection{Global Contraction of the Transport Map}
\label{sec:global_geometry}

While our bounds link single-step viability to posterior compactness, evaluating target dispersion directly from offline demonstrations is impossible, as each context $c = (o_t, l, s_t)$ typically admits only a single trajectory. A trained flow model bridges this gap: by providing a smooth generative map $x_1 = \phi_1^\theta(x_0; c)$ from prior noise to actions, its Jacobian $J(x_0; c) \triangleq \nabla_{x_0} \phi_1^\theta(x_0; c)$ acts as an exact differential probe of how conditioning contracts the target measure.

To capture this behavior across the entire action manifold, we evaluate the empirical population distribution of the spectral norm $\sigma_1(J(x_0; c)) = \|\nabla_{x_0}\phi_1^\theta\|_2$ by jointly sampling condition contexts from evaluation demonstrations and prior noise $x_0 \sim \mathcal{N}(0, I)$.
Crucially, rather than characterizing an isolated context, this empirical distribution serves as a dataset-level diagnostic of typical transport geometry across tasks.
To probe the true transport geometry, we evaluate $J(x_0; c) = \nabla_{x_0}\phi_1^\theta$ where $\phi_1^\theta(x_0; c) \triangleq x_0 + \int_0^1 v_\theta(x_t, t, c)\,dt$ for flow-matching model, and $\phi_1^\theta(x_0; c) \triangleq G_\theta(x_0, c)$ for native one-step models (MeanFlow~\citep{geng2025meanflow}, pMF~\citep{lu2026onesteplatentfreeimagegeneration}).

\begin{figure}[!htbp]
\centering
\includegraphics[width=0.88\linewidth]{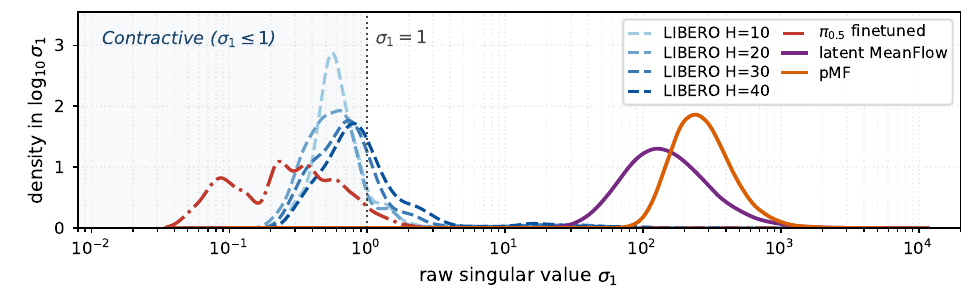}
\caption{\textbf{Empirical distribution of the generative Jacobian spectral norm $\sigma_1(J)$.} 
KDE densities over $\log_{10}\sigma_1$ across domains. Output conventions cannot account for this gap, as variance-matching the $[-1,1]$ image coordinates would only enlarge the image spectra (Appendix~\ref{app:sigma_normalization}).}
\label{fig:sigma1_global}
\end{figure}

As shown in Figure~\ref{fig:sigma1_global}, the cross-domain divide is unequivocal: singular values for VLA policies concentrate predominantly below the contractive threshold $\sigma_1 = 1$, a predominantly contractive regime consistent with the concentration behavior of \cref{eq:subgaussian_norm}, whose exponential guarantee itself rests on the stronger uniform bound $L(c)$.
In contrast, image flows are strictly expansive, operating two to three orders of magnitude higher.
Within the VLA family, extending the action horizon progressively shifts mass past $\sigma_1 = 1$, capturing the loss of contractivity that underpins the performance degradation in Section~\ref{sec:action_horizon}.

\subsection{Local Geometry: $\sigma_1$ Fractures Trace Decision Boundaries}
\label{sec:local_geometry}

Beyond aggregate contractivity (Section~\ref{sec:global_geometry}), $\sigma_1(x_0; c)$ measures local sensitivity to initial noise perturbations.
Mapping a smooth Gaussian prior onto separated data modes forces the transport map to stretch at the interfaces, which manifest as high-$\sigma_1$ ridges separating outcome basins.
Throughout this section we write the Jacobian SVD as $J = \sum_i \sigma_i\, u_i v_i^\top$, with $v_i$ acting in noise space and $u_i$ in action space.
We probe this structure identically across domains: fix the condition, sample Gaussian noise vectors, screen for high-$\sigma_1$ anchors, and map $\sigma_1$ on the plane spanned by each anchor's leading right singular directions $(v_1, v_2)$, decoding along a circular orbit that crosses the ridge.

\begin{figure}[!htbp]
\centering
\includegraphics[height=4.4cm]{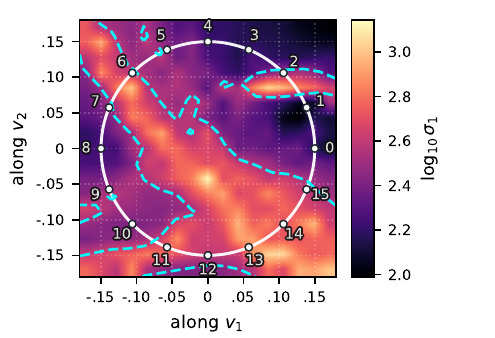}\hfill
\includegraphics[height=4.4cm]{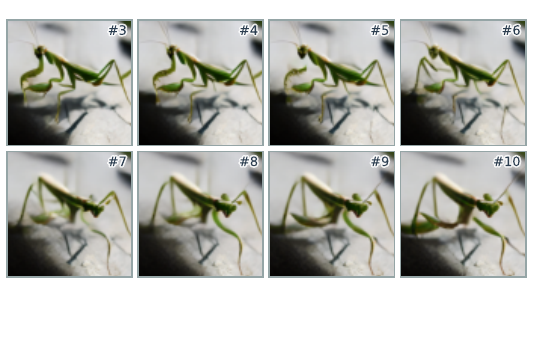}
\caption{\textbf{Local transport geometry of an image flow.}
(a) $\log_{10}\sigma_1$ over the $(v_1, v_2)$ noise plane; the circular orbit (cyan) crosses the p95 ridge.
(b) Generations along the orbit; crossing the ridge flips the pose between \#6 and \#7.}
\label{fig:local_geometry_mantis}
\vspace{-0.7em}
\end{figure}

We first inspect visual synthesis, where mode boundaries are semantic transitions.
Fixing a class label of the MeanFlow model, a prominent singular-value ridge emerges in the $(v_1, v_2)$ plane (Figure~\ref{fig:local_geometry_mantis}a); crossing it on the orbit abruptly flips the mantis pose between samples \#6 and \#7, while poses on either side remain stable (Figure~\ref{fig:local_geometry_mantis}b).
The flow thus maps mode adjacency back into noise space, where $\sigma_1$ exposes the interfaces between coherent semantic modes.

\begin{figure}[!htbp]
\centering
\subcaptionbox{episode $\sigma_1$ trace}{\includegraphics[height=2.3cm]{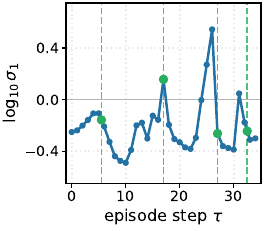}}%
\hfill
\subcaptionbox{$\log_{10}\sigma_1$ on $(v_1,v_2)$}{\includegraphics[height=2.3cm]{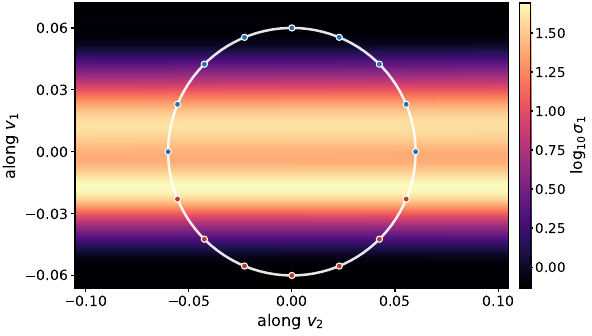}}%
\hfill
\subcaptionbox{orbit decodes}{\includegraphics[height=2.3cm]{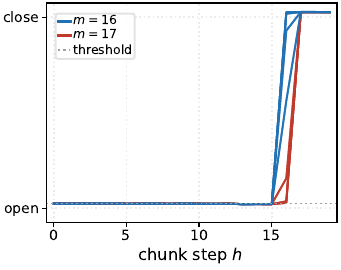}}%
\hfill
\subcaptionbox{$h^*$ vs.\ recorded}{\includegraphics[height=2.3cm]{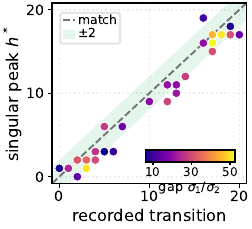}}
\vspace{-0.4em}
\caption{\textbf{Action flow geometry during manipulation.} Protocols and additional panels: Appendix~\ref{app:geometry}.}
\label{fig:action_local_geometry}\label{fig:causal_timing_mechanism}\label{fig:local_geometry_libero}\label{fig:libero_transport_geometry}
\vspace{-1.0em}
\end{figure}

In flow policies, we sample context--noise pairs jointly: contexts $c = (o_t, l, s_t)$ from closed-loop episodes, $x_0\sim\mathcal{N}(0,I)$, screened by $\sigma_1(x_0;c)$.
The high-$\sigma_1$ pairs cluster near \emph{gripper transitions}---steps at which the decoded chunk switches the gripper---while $\sigma_1$ stays below $1$ almost everywhere else (Fig.~\ref{fig:action_local_geometry}a), keeping the rollout concentrated around the conditional mean (\cref{eq:subgaussian_norm}; protocols in Appendix~\ref{app:geometry}).

Write $m(x_0;c)\in\{1,\dots,H\}$ for the first chunk step of $\phi_1^\theta(x_0;c)$ in the new gripper state.
At an anchor near such a transition, a sharp $\sigma_1$ ridge emerges in the $(v_1, v_2)$ plane, and crossing it shifts $m$ exactly once (Fig.~\ref{fig:action_local_geometry}b,c).

Two checks establish this ridge as a physical decision boundary:
(i)~\textbf{Causal specificity:} Pushing initial noise along $v_1$ ($\epsilon \le 0.1$) reliably shifts the switch step of screened events in the top $\sigma_1$ decile (by at most two chunk steps), whereas a pure-noise policy responds in only 6.5\% of cases (Appendix~\ref{app:geometry}, Fig.~\ref{fig:app_local_mechanism}a).
(ii)~\textbf{Predictive timing:} The gripper-channel singular peak $h^* \triangleq \operatorname{argmax}_h (u_1^{\mathrm{grip}}[h])^2$ falls within two chunk steps of the \emph{recorded} gripper transition at 81\% of anchors (median offset $-1$; Fig.~\ref{fig:action_local_geometry}d).

Together, these views reconcile: global contractivity concentrates one-step samples near the conditional mean via Gaussian concentration (\cref{eq:subgaussian_norm}), while spectral fractures isolate discrete physical decisions.
Even under single-step decoding, this boundary timing signature is preserved far more accurately than under pure-noise training (Appendix~\ref{app:onestep_geometry}).

\paragraph{Condition modalities and geometry.}
Unlike extending the action horizon, condition modalities exhibit structural redundancy and regularization rather than a simple monotonic ordering (geometric analysis in Appendix~\ref{sec:condition_geometry}).

\section{Discussion}
\enlargethispage{2\baselineskip}

Across diverse manipulation benchmarks and physical bimanual tasks, rich conditioning over a compact target enables single-step flow matching without distillation. The velocity-loss profile diagnoses this structure from the Bayesian side; Gaussian concentration (\cref{eq:subgaussian_norm}) explains it from the geometric side---a contractive transport map ($\sigma_1<1$ almost everywhere) renders the conditional target sub-Gaussian about its mean, while expansive image flows sit in the opposite regime. Yet compact is not collapsed: residual structure survives as sharp spectral fractures along physical decision boundaries, which flow matching causally encodes and neither direct regression nor pure-noise training resolves.

This shifts the scaling frontier from action decoding to condition construction. An agentic VLM actively enriches its own context through reasoning and interaction; because the resulting action target remains compact, decoding stays trivially cheap (Appendix~\ref{app:latency}), rendering elaborate action-space diffusion machinery secondary to condition representation.


\newpage
\section*{AI use statement}

In this work, we used generative AI tools for tasks with recommended disclosure: creating and editing software code (analysis and plotting scripts), creating and modifying scientific figures, and editing the manuscript to improve readability and grammar. We have not used generative AI tools for tasks with required disclosure; such tasks are not applicable to this work or were carried out by the authors. We have reviewed all AI-assisted work: AI-generated code was verified against experiment logs and outputs by the authors, and all AI-assisted figures and text were checked and approved by the authors. We take responsibility for the final content of this work, including text, claims, or artifacts produced with the aid of generative AI.

\bibliography{references}
\bibliographystyle{conference_style}

\newpage
\appendix
\section{Proof of \Cref{eq:subgaussian_norm}}
\label{app:proof_subgaussian}

\begin{theorem}[Gaussian transport concentration]
\label{thm:subgaussian_transport}
Fix a condition $c$, let $z\sim\mathcal N(0,I_d)$, and define $x_1=\phi_1(z;c)$,
where $\phi_1(\cdot;c)\in W^{1,\infty}_{\mathrm{loc}}(\mathbb R^d;\mathbb R^d)$
has Jacobian $J(z;c)=\nabla_z\phi_1(z;c)$. Let
\[
  L(c)\triangleq\operatorname*{ess\,sup}_{z}\|J(z;c)\|_2<\infty .
\]
Then every unit projection $u^\top x_1$, $u\in\mathbb S^{d-1}$, is
$L(c)^2$-sub-Gaussian:
\begin{equation}
  \mathbb E_z\!\left[\exp\!\Big(\lambda\,u^\top\big(x_1-\mathbb E_z[x_1]\big)\Big)\right]
  \le
  \exp\!\left(\frac{\lambda^2L(c)^2}{2}\right),
  \qquad \forall\,\lambda\in\mathbb R .
\end{equation}
Moreover,
\begin{equation}
\label{eq:subgaussian_norm_app}
  \mathbb P_z\!\left(
    \|x_1-\mathbb E_z[x_1]\|_2
    \ge
    \sqrt{\mathbb E_z\|J(z;c)\|_F^2}+\epsilon
  \right)
  \le
  \exp\!\left(-\frac{\epsilon^2}{2L(c)^2}\right),
  \qquad \forall\,\epsilon>0 .
\end{equation}
Since $\mathbb E_z\|J(z;c)\|_F^2\le d\,L(c)^2$, the characteristic radius is
further upper bounded by $L(c)\sqrt d$.
\end{theorem}

\begin{proof}
Since $\mathbb R^d$ is convex and $\|J\|_2\le L(c)$ almost everywhere,
$\phi_1(\cdot;c)$ is globally $L(c)$-Lipschitz. In particular, all
expectations below are finite.

\medskip
\noindent\emph{Projections.}
For $u\in\mathbb S^{d-1}$, let $g(z)=u^\top\phi_1(z;c)$. Then
\[
  \|\nabla g(z)\|_2=\|J(z;c)^\top u\|_2\le L(c)
  \quad\text{almost everywhere.}
\]
The standard Gaussian measure $\gamma$ satisfies the logarithmic Sobolev
inequality
\[
  \operatorname{Ent}_\gamma(f^2)\le 2\,\mathbb E_\gamma\big[\|\nabla f\|_2^2\big].
\]
Applying it with $f=e^{\lambda g/2}$ and using the Herbst argument yields
\begin{equation}
  \mathbb E_z\!\left[\exp\!\big(\lambda(g-\mathbb E_z[g])\big)\right]
  \le
  \exp\!\left(\frac{\lambda^2L(c)^2}{2}\right)
\end{equation}
for $\lambda>0$; applying the same argument to $-g$ gives the result for all
$\lambda\in\mathbb R$.

\medskip
\noindent\emph{Norm.}
Define
\[
  F(z)=\big\|\phi_1(z;c)-\mathbb E_z[\phi_1(z;c)]\big\|_2 .
\]
Since $\phi_1(\cdot;c)$ is $L(c)$-Lipschitz, the reverse triangle inequality
shows that $F$ is $L(c)$-Lipschitz. Hence Gaussian concentration gives
\begin{equation}
\label{eq:norm_concentration}
  \mathbb P_z\!\left(F\ge\mathbb E_z[F]+\epsilon\right)
  \le
  \exp\!\left(-\frac{\epsilon^2}{2L(c)^2}\right).
\end{equation}
Next, since $\mathbb E_z[F^2]=\sum_{i=1}^d\operatorname{Var}_z\big(\phi_1^{(i)}\big)$,
the Gaussian Poincar\'e inequality applied coordinate-wise gives
\begin{equation}
  \mathbb E_z[F^2]
  =
  \sum_{i=1}^d\operatorname{Var}_z\!\left(\phi_1^{(i)}(z;c)\right)
  \le
  \sum_{i=1}^d\mathbb E_z\!\left[\big\|\nabla_z\phi_1^{(i)}(z;c)\big\|_2^2\right]
  =
  \mathbb E_z\|J(z;c)\|_F^2 .
\end{equation}
Therefore, by Jensen's inequality,
\[
  \mathbb E_z[F]\le\sqrt{\mathbb E_z[F^2]}\le\sqrt{\mathbb E_z\|J(z;c)\|_F^2}.
\]
Consequently,
\[
  \left\{F\ge\sqrt{\mathbb E_z\|J(z;c)\|_F^2}+\epsilon\right\}
  \subseteq
  \left\{F\ge\mathbb E_z[F]+\epsilon\right\},
\]
and \cref{eq:subgaussian_norm} follows from \cref{eq:norm_concentration}.

\medskip
\noindent\emph{Radius bound.}
Since $J(z;c)$ is $d\times d$, $\|J(z;c)\|_F^2\le d\,\|J(z;c)\|_2^2\le d\,L(c)^2$
almost everywhere, which yields the stated coarse radius bound.
\end{proof}

\section{Time Coordinates and Training Schedules}
\label{app:time}

The paper uses the flow-matching coordinate \(t\), with \(t=0\) denoting the
noise endpoint and \(t=1\) denoting clean data. OpenPI uses the opposite
coordinate \(t_{\mathrm{op}}=1-t\), where \(t_{\mathrm{op}}=1\) is pure noise.
Its default base distribution \(t_{\mathrm{op}}\sim\mathrm{Beta}(1.5,1)\)
therefore becomes \(t\sim\mathrm{Beta}(1,1.5)\) in the paper coordinate.
For an OpenPI-coordinate sample, the high-noise shift is
\(t_{\mathrm{op,shifted}} = \alpha t_{\mathrm{op}} / (1+(\alpha-1)t_{\mathrm{op}})\).
Larger \(\alpha\) moves probability mass toward
\(t_{\mathrm{op}}=1\), equivalently \(t=0\), while preserving the
velocity-prediction target. This is the same schedule as in
Section~\ref{sec:libero_schedule} after changing coordinates.

All diagnostics are plotted in the paper coordinate (OpenPI logs converted as
\(t=1-t_{\mathrm{op}}\)).

\begin{wraptable}{r}{0.55\linewidth}
\vspace{-1.4em}
\centering
\caption{\textbf{Direct learned action-start controls on standard LIBERO} (success rates \%). All rows use the Tiny H10 policy with one-step decoding (500 episodes per suite). Differences ($\pm$) are relative to the pure-noise 1-step Tiny baseline (Table~\ref{tab:time_ablation}).}
\label{tab:learned_start}
\footnotesize
\setlength{\tabcolsep}{2pt}
\begin{tabular}{lcccc}
\toprule
\textbf{Learned Start Init} & \textbf{Spatial} & \textbf{Object} & \textbf{Goal} & \textbf{Long} \\
\midrule
Zero            & 92.8 \negdiff{3.2} & 99.0 \posdiff{1.2} & 59.4 \negdiff{36.8} & 68.2 \negdiff{4.6} \\
Normal, std = 1 & 95.4 \neudiff{-0.6} & 98.6 \posdiff{0.8} & 59.8 \negdiff{36.4} & 69.4 \negdiff{3.4} \\
\bottomrule
\end{tabular}
\vspace{-0.8em}
\end{wraptable}

\begin{figure}[!htbp]
\centering
\includegraphics[width=0.95\linewidth]{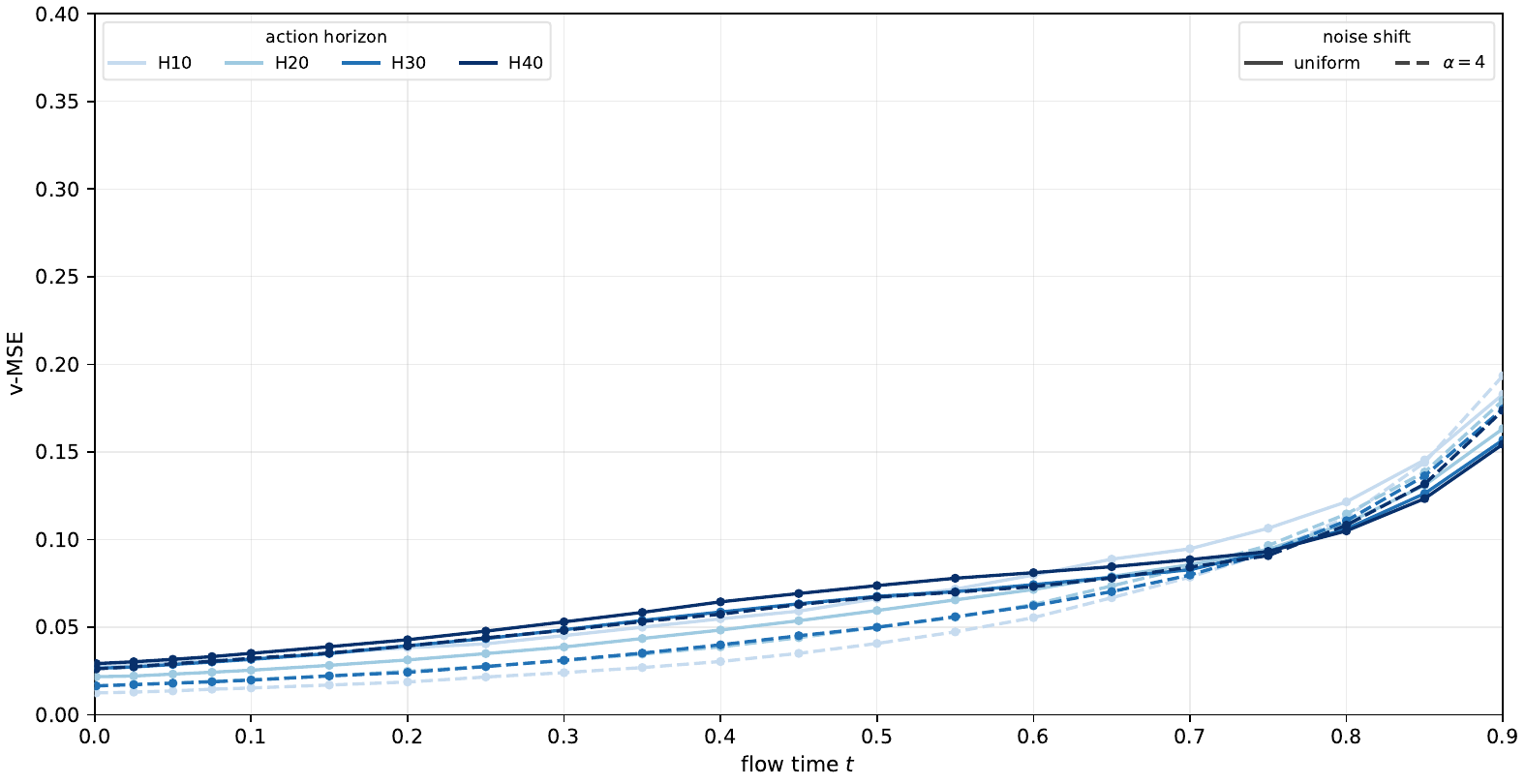}
\caption{Full horizon sweep underlying Fig.~\ref{fig:horizon_vloss}, including the H10 curves. Color denotes the horizon (H10 light to H40 dark); linestyle denotes the training-time distribution (uniform solid, $\alpha{=}4$ dashed).}
\label{fig:horizon_vloss_full}
\vspace{-0.6em}
\end{figure}

\paragraph{Velocity-MSE protocol.}
All velocity-MSE profiles (Figs.~\ref{fig:vloss_profiles}, \ref{fig:horizon_vloss}, and
\ref{fig:horizon_vloss_full}) are computed by the same fixed-time diagnostic.
For each checkpoint we sample a fixed set of evaluation contexts and, for each
context, draw a matched Gaussian \(x_0\sim\mathcal{N}(0,I)\); the matched-noise
anchor protocol draws a single long noise sequence and slices prefixes of length
\(H\), so every horizon sees identical noise. We then evaluate the model at 25
paper times \(t\in[0.001,0.999]\) and regress the predicted velocity against the
analytic target \(u=x_1-x_0\), reporting the mean squared error over the seven
physical action dimensions (the 32-dimensional action tensor is sliced to its
first seven coordinates; the padding coordinates never enter any loss or
metric). Each profile aggregates 64 context--noise pairs.

\section{Transport-Geometry Diagnostics}
\label{app:geometry}

This appendix specifies how every \(\sigma_1\) quantity in
Section~\ref{sec:geometry} is computed. Unless stated otherwise, all estimators share the same object:
the noise-to-action map \(\phi_1^\theta(\cdot\,;c)\) and its Jacobian
\(J(x_0;c)=\nabla_{x_0}\phi_1^\theta(x_0;c)\).

\subsection{Protocols}

\paragraph{Endpoint Jacobian.}
For models that natively parameterize one-step transport (MeanFlow, pMF) we
differentiate the single forward map directly. For continuous flow models
(\(\pi_0\), \(\pi_{0.5}\), StarVLA, X-VLA, and our LIBERO checkpoints) the
trained vector field \(v_\theta(x,t,c)\) is integrated along the paper-time
path from \(t=0\) to \(t=1\) with 10-step Euler integration, and
\(\phi_1^\theta\) denotes the composed map; \(J\) is obtained by reverse-mode
autodifferentiation through the composition. At each evaluation point we form
the Jacobian and take its largest singular value, either by a direct SVD for
small action dimensions or by power iteration on \(J^{\top}J\).

\paragraph{Transition screening.}
Gripper transitions are screened from held-out rollouts of the H20 uniform-schedule policy: around each recorded gripper close we draw a dense candidate set of 444 noise sources, evaluate the \(\phi_1^\theta\) spectrum of each, and retain high-\(\sigma_1\) states with an observable gripper switch at chunk step \(m\); greedy no-replacement selection across episodes yields 32 events.

\begin{figure}[!htb]
\centering
\subcaptionbox{causal steering ($v_1$)}{\includegraphics[height=2.9cm]{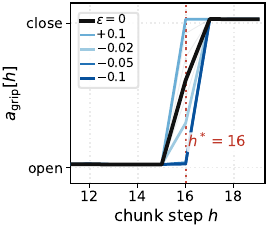}}%
\hfill
\subcaptionbox{actuation trace}{\includegraphics[height=2.9cm]{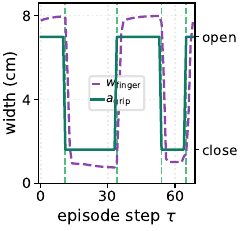}}%
\hfill
\subcaptionbox{core cross-section}{\includegraphics[height=2.9cm]{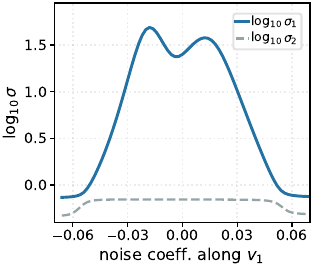}}%
\hfill
\subcaptionbox{decoded switch step}{\includegraphics[height=2.9cm]{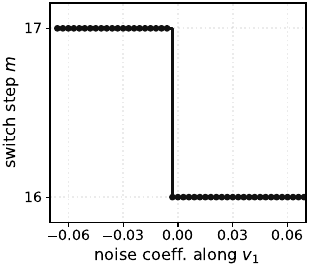}}
\vspace{-0.2em}
\caption{\textbf{Steering scans and landscape cross-sections.}
(a) Commanded gripper action $a_{\text{grip}}$ along chunk steps $h$ under initial noise shifts $x_0 + \epsilon v_1$.
(b) Commanded gripper action $a_{\text{grip}}$ and physical finger width $w_{\text{finger}}$ confirm that fractures coincide with physical contact transitions.
(c) Cross-section at $b{=}0$ of the $\phi_1^\theta$ landscape in Fig.~\ref{fig:libero_transport_geometry}: $\sigma_1$ (blue) forms two lanes flanking a saddle, while $\sigma_2$ (gray dashed) stays below $0.7$.
(d) Decoded switch step $m(a)$ at $b{=}0$: a single jump inside the saddle region.}
\label{fig:app_local_mechanism}
\vspace{-0.6em}
\end{figure}

\paragraph{Steering scans.}
Each event state is re-decoded under shifted noise \(x_0+\epsilon v_1\), with \(\epsilon\) on the symmetric grid \(\{0,\pm0.02,\pm0.05,\pm0.1,\pm0.2,\pm0.5\}\); the switch step \(m\) is read off the decoded gripper trace, and an event counts as switched when \(m(+\epsilon)\neq m(-\epsilon)\) at some paired magnitude (Fig.~\ref{fig:app_local_mechanism}a). The main text reports only \(\epsilon\le0.1\), where shifted probes remain in the typical Gaussian region.

\paragraph{Pure-noise steering control.}
The pure-noise policy is natively one-step and has no multi-step flow to compose, so its control probe targets its own one-step map: the H20 pure-noise checkpoint is screened by the same high-\(\sigma_1\) procedure on that map and re-decoded under the same \(\epsilon\) grid and switch statistic as above.

\paragraph{Landscape cross-sections.}
For one transition (episode 18, step 5116; H20 uniform-schedule policy) we evaluate the fp32 flow map \(\phi_1^\theta\) (10-step Euler composition) on a \(70\times81\) grid in the \((v_1, v_2)\) plane, producing the landscape of Fig.~\ref{fig:action_local_geometry}b; a circular orbit of radius \(0.06\) and a 1D decode lattice along \(a\) at \(b{=}0\) produce the orbit decodes (Fig.~\ref{fig:action_local_geometry}c), the \(\sigma\) cross-section, and the quantized switch step \(m(a)\) (Fig.~\ref{fig:app_local_mechanism}c,d).

\paragraph{Episode-time traces.}
Figure~\ref{fig:local_geometry_libero}a follows the H20 uniform-schedule policy along one held-out rollout segment, sampling an observation every 8 frames around its recorded gripper boundaries (green markers). At each observation \(c_\tau\), conditioned on the recorded frame, \(\sigma_1\) of the complete-map Jacobian \(\nabla_z\phi_1^\theta(\,\cdot\,;c_\tau)\)---the same 10-step Euler map as in the landscape protocol above---is estimated matrix-free by block power iteration (block rank 2, three subspace iterations) at \(K=4\) fixed Gaussian noise sources, and the mean \(\log_{10}\sigma_1\) over sources is plotted. The trace stays flat and contractive between boundaries, climbs across \(\sigma_1{=}1\) ahead of each boundary, and drops back immediately after; the commanded gripper action and physical finger width around such transitions are shown in Fig.~\ref{fig:app_local_mechanism}b.

\paragraph{Noise-plane heatmaps.}
Figure~\ref{fig:local_geometry_mantis}a evaluates a latent MeanFlow model on the
plane \(z_0+a v_1+b v_2\) around a fixed anchor \(z_0\); \(\sigma_1\) is
evaluated on a \(17\times17\) grid by per-point autodiff, and the orbit
\(z(\alpha)\) is the circle of radius \(0.15\) in this plane. The cyan contour
marks the p95 ridge of the grid spectrum.

\paragraph{Population distributions.}
The histograms of Figs.~\ref{fig:sigma1_global} and~\ref{fig:condition_sigma1}
are built by sampling evaluation contexts \(c=(o_t,l,s_t)\) from held-out
demonstrations together with i.i.d.\ Gaussian noise \(x_0\), evaluating
\(\sigma_1(x_0;c)\) at each pair (1088 pairs per condition cell), and plotting
the density in \(\log_{10}\sigma_1\). Fig.~\ref{fig:sigma1_global} uses the
endpoint Jacobian; Fig.~\ref{fig:condition_sigma1} uses \(\phi_1^\theta\)
(\(10\)-step Euler) of the H10 \(\alpha{=}4\) policy.

\begin{figure}[!htbp]
\centering
\includegraphics[width=0.9\linewidth]{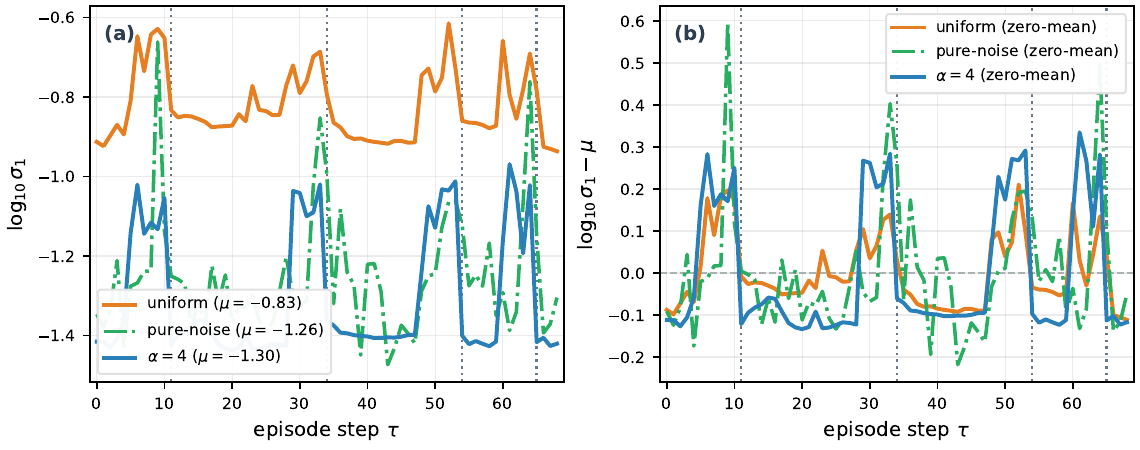}
\caption{Episode geometry across training schedules. (a) Per-step \(\log_{10}\sigma_1\) of the same episode under uniform, pure-noise, and \(\alpha{=}4\) training (means \(-0.83\), \(-1.26\), \(-1.30\)). (b) Zero-mean relative sensitivities around the same boundaries.}
\label{fig:app_noise_invariance}
\vspace{-0.6em}
\end{figure}

\paragraph{Schedule specificity.}
Figure~\ref{fig:app_noise_invariance} repeats the per-step diagnostic on the same episode under the three training schedules, now with the one-step generator map \(T_c(z)=z+v_\theta(z,t{=}0;c)\) (\(K{=}16\) fixed noise anchors, full SVD) so that the natively one-step pure-noise policy enters on equal footing. The quiet-between-transitions structure is specific to the flow map: the pure-noise trace stays noisy throughout.
The same specificity holds under intervention: at matched small shifts (\(\epsilon\le0.1\)) the \(v_1\)-steering probe of Section~\ref{sec:local_geometry} changes \(m\) in only 6.5\% of screened pure-noise events (18.3\% at \(\epsilon\le0.2\); 34.4\% over the full grid), whereas \(\phi_1^\theta\) responds reliably already at \(\epsilon\le0.1\).

\subsection{Boundary Timing of the Deployed One-Step Map}
\label{app:onestep_geometry}

\begin{figure}[!htbp]
\centering
\includegraphics[width=0.52\linewidth]{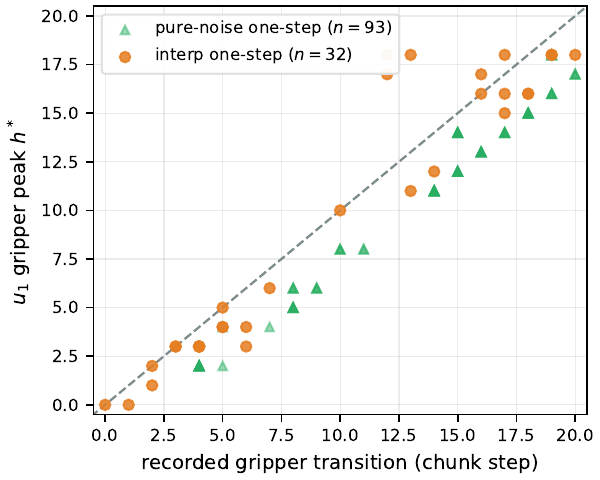}
\caption{\textbf{The deployed one-step map retains boundary timing.} Gripper-channel peak $h^*$ of the one-step map's leading output direction vs.\ the recorded gripper transition step, at the 32 screened anchors (H20 uniform): the interpolation-trained policy tracks the physical timing, while the pure-noise control (screened at its own anchors) peaks systematically early. Dashed: exact match.}
\label{fig:app_onestep_geometry}
\vspace{-0.6em}
\end{figure}

Section~\ref{sec:local_geometry} probes the composed flow map $\phi_1^\theta$ (10-step Euler), while deployment decodes by a single vector-field evaluation $T_c(z) = z + v_\theta(z, t{=}0; c)$.
The deployed map still carries the boundary's timing: at the same 32 screened anchors (H20 uniform), $h^*$ falls within two chunk steps of the recorded gripper transition for 88\% of anchors, whereas the pure-noise map's $h^*$ peaks systematically three steps early (28\%; Fig.~\ref{fig:app_onestep_geometry}).

\subsection{Output Normalization and Cross-Domain Spectra}
\label{app:sigma_normalization}

The spectral norm $\sigma_1(J)$ is coordinate dependent, and the two domains in Fig.~\ref{fig:sigma1_global} use different output conventions: action coordinates are per-coordinate standardized with training-set mean and standard deviation (unit marginal variance), whereas image models use the conventional pixel scaling $x_i\in[-1,1]$. The following bound shows this mismatch cannot explain the cross-domain gap: variance-matching the image coordinates would only enlarge the image spectra.

\begin{proposition}[Coordinate-wise standardization]
\label{prop:normalization_sigma}
Let $D=\operatorname{diag}(s_1,\dots,s_d)$, $s_i>0$, and $\widetilde x = D^{-1}(x-\mu)$ where $x=\phi_1^\theta(z;c)$. Translation does not affect the Jacobian, so $J_{\widetilde x}=D^{-1}J$, and
\[
\frac{\sigma_1(J)}{s_{\max}}\;\le\;\sigma_1(D^{-1}J)\;\le\;\frac{\sigma_1(J)}{s_{\min}},
\qquad s_{\min}=\min_i s_i,\; s_{\max}=\max_i s_i .
\]
\end{proposition}

\begin{proof}
For any unit vector $v$, $\|D^{-1}Jv\|_2 \ge \sigma_{\min}(D^{-1})\,\|Jv\|_2 = \|Jv\|_2/s_{\max}$ and $\|D^{-1}Jv\|_2 \le \|D^{-1}\|_2\,\|Jv\|_2 = \|Jv\|_2/s_{\min}$; taking the supremum over $\|v\|_2=1$ gives both bounds.
\end{proof}

Every image coordinate is supported on $[-1,1]$, hence its marginal standard deviation satisfies $s_i^{\mathrm{img}}\le1$. Standardizing image outputs to unit marginal variance therefore gives
$\sigma_1(J_{\mathrm{img,std}}) \ge \sigma_1(J_{\mathrm{img}})/s_{\max}^{\mathrm{img}} \ge \sigma_1(J_{\mathrm{img}})$:
the reported comparison is conservative with respect to this scale mismatch. This statement concerns only coordinate-wise scaling---$\sigma_1$ is not claimed invariant under general linear reparameterizations such as full covariance whitening---and all measurements are taken in each model's native training coordinates.

\section{Condition Ablations and Closed-Loop Proprioceptive Failure}
\label{sec:condition_geometry}
\label{app:condition_ablations}

We ablate individual condition components of the H10 $\alpha=4$ one-step policy (Table~\ref{tab:condition_controls}, 500 episodes per suite), and the degradations are highly non-uniform.
Visual ablation leaves basic reaching intact on simpler suites---the Object suite swaps only the target object at fixed layouts, so it remains solvable from proprioception and task context---but collapses long-horizon tasks (Long drops to 31.8\% and 53.2\%).
Language ablation causes an isolated semantic collapse on Goal (8.8\%), where scenes are identical across tasks and the prompt is the sole grounding signal.
Proprioceptive ablation causes widespread prehensile failure (Object 18.2\%, Long 2.0\%) while Goal largely survives (44.0\%).

\begin{figure}[!htbp]
\centering
\includegraphics[width=0.78\linewidth]{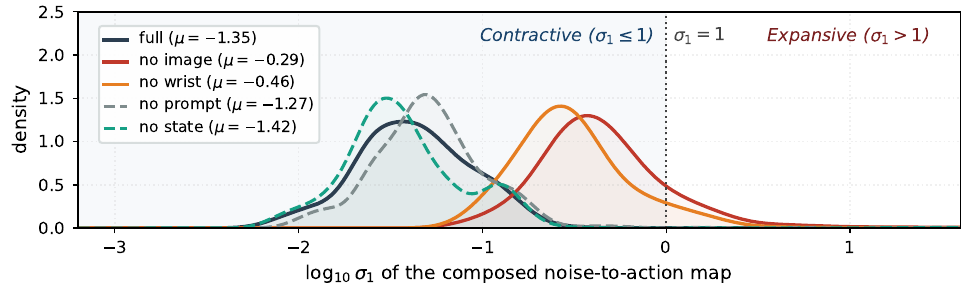}
\caption{\textbf{Transport geometry under condition ablations.} Distribution of $\log_{10}\sigma_1$ of the noise-to-action map for the H10 $\alpha{=}4$ policy (Table~\ref{tab:condition_controls}): removing the image or the wrist camera pushes mass across the contractive boundary $\sigma_1{=}1$ (dotted), i.e., a genuinely broadened conditional distribution; removing the prompt or the state leaves the geometry essentially unchanged---prompt deletion in particular induces categorical semantic ambiguity rather than continuous geometric dispersion.}
\label{fig:condition_sigma1}
\vspace{-0.6em}
\end{figure}

\begin{table}[!htbp]
\centering
\caption{Condition ablations on standard LIBERO (H10 $\alpha{=}4$ one-step tiny policy, success rates \%, 500 episodes per suite).}
\label{tab:condition_controls}
\small
\setlength{\tabcolsep}{7pt}
\begin{tabular}{lrrrr}
\toprule
Condition & Spatial & Object & Goal & Long \\
\midrule
Full & 96.4 & 99.6 & 96.8 & 85.2 \\
No image & 82.2 & 96.0 & 85.4 & 31.8 \\
No prompt & 82.0 & 97.2 & 8.8 & 55.2 \\
No wrist & 92.2 & 96.4 & 88.0 & 53.2 \\
No state & 33.8 & 18.2 & 44.0 & 2.0 \\
\bottomrule
\end{tabular}
\end{table}

\paragraph{Closed-loop failure of the no-state policy.}
The no-state ablation leaves the transport geometry essentially unchanged---$\sigma_1$ even contracts slightly (Fig.~\ref{fig:condition_sigma1})---and the policy tracks recorded demonstrations as accurately as the full policy in open loop (per-step end-effector error ratio 0.93--1.03): LIBERO's fixed scenes make the arm state largely estimable from vision. What fails is closed-loop control, and a $2{\times}2$ same-root intervention (Table~\ref{tab:same_root}) shows the state channel itself is causal: restarting from exact environment snapshots of failed no-state rollouts---taken just before or at the grasp---the full policy recovers 94.4\% and 91.7\% of them respectively, the no-state policy itself only 5.6\%. Our tentative reading: the state channel matters less as extra information than as making the condition structure explicit and easy to learn, which is what generalizes off the demonstrated trajectory.

\begin{table}[!htbp]
\centering
\caption{\textbf{Same-root intervention} (success rates \%). Paired snapshots of successful full-policy and failed no-state rollouts (LIBERO Object) are taken at matched timesteps before the grasp (approach) and at the grasp (event), then continued under both policies; full/no-state columns denote which rollout supplied the snapshot. 36 continuations per cell (18 snapshots $\times$ 2 noise sources).}
\label{tab:same_root}
\small
\setlength{\tabcolsep}{7pt}
\begin{tabular}{lcccc}
\toprule
& \multicolumn{2}{c}{Approach roots} & \multicolumn{2}{c}{Event roots} \\
\cmidrule(lr){2-3} \cmidrule(l){4-5}
Policy & full & no-state & full & no-state \\
\midrule
Full & 100.0 & 94.4 & 100.0 & 91.7 \\
No-state & 38.9 & 5.6 & 50.0 & 5.6 \\
\bottomrule
\end{tabular}
\end{table}

\section{Image-to-Text MNIST Details}
\label{app:mnist}

The condition is a $128\times128$ image of a $4\times4$ grid of MNIST digits~\citep{lecun1998mnist}, and the target is the corresponding 16-token digit sequence, read in row-major order. A CNN encoder embeds the grid image into condition tokens. The target token embeddings $x_1 \in \mathbb{R}^{16\times256}$ follow the standard linear interpolation $x_t = t x_1 + (1-t) x_0$ with $x_0 \sim \mathcal{N}(0,I)$. The noised target tokens and the condition tokens are concatenated and fed into a bidirectional Transformer decoder without causal masking, which predicts the velocity $v_\theta(x_t, t, c)$.

\begin{figure}[!htbp]
    \centering
    \begin{minipage}[t]{0.46\linewidth}
        \centering
        \includegraphics[width=\linewidth]{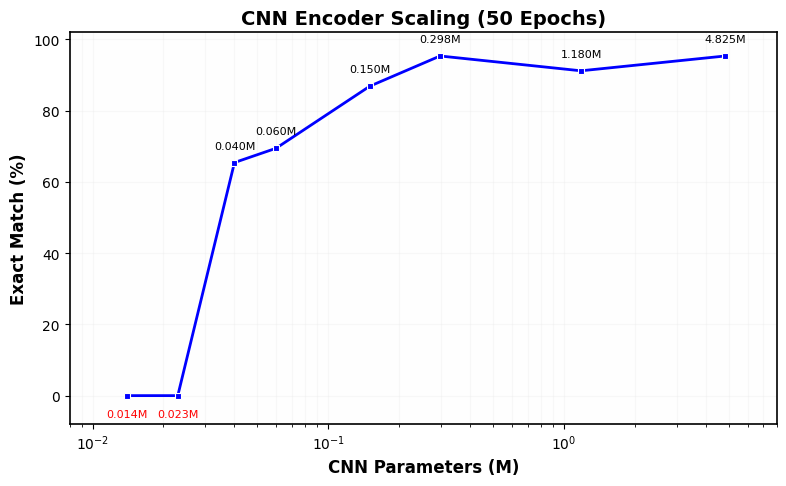}
        {\footnotesize (a) CNN encoder scale.}
    \end{minipage}\hfill
    \begin{minipage}[t]{0.46\linewidth}
        \centering
        \includegraphics[width=\linewidth]{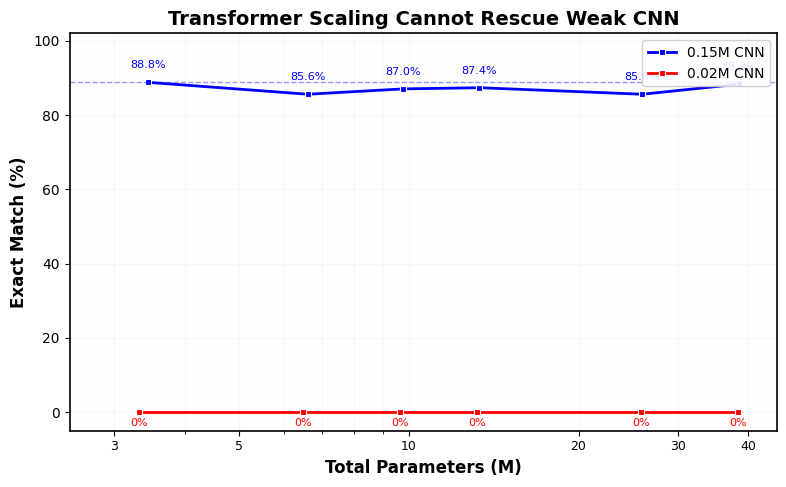}
        {\footnotesize (b) Decoder scale compensation.}
    \end{minipage}
    \caption{Two ablations: (a) varying the CNN encoder scale shows that weaker encoders degrade one-step performance; (b) scaling up the Transformer decoder while keeping a weak encoder does not compensate, confirming that the bottleneck lies in the condition-target structure, not model capacity.}
    \label{fig:mnist_ablations}
\end{figure}

The image-to-text MNIST dataset contains 3,750 training examples and 625 test
examples. In addition to the flow-matching loss, we use a cross-entropy term
on the clean-target prediction,
$
    \mathcal{L}
    = \mathcal{L}_{\mathrm{FM}}+\lambda\mathcal{L}_{\mathrm{CE}},\lambda=2,$
where \(x_{\mathrm{pred}}=x_t+(1-t)v_{\mathrm{pred}}\) is used to evaluate
\(\mathcal{L}_{\mathrm{CE}}\). Table~\ref{tab:app_mnist_hparams} gives the
shared training configuration behind the MNIST schedule and encoder/decoder
controls in the main text.

\begin{table}[H]
\centering
\caption{Image-to-text MNIST training hyperparameters.}
\label{tab:app_mnist_hparams}
\small
\begin{tabular}{@{}ll@{}}\toprule Parameter & Value \\ \midrule
Image size & \(128\times128\) \\
Maximum text length & 20 tokens \\
Vocabulary size & 13 \\
\(d_{\mathrm{model}}\) / layers / heads & 256 / 4 / 8 \\
Feedforward dimension & 1024 \\
Dropout & 0.1 \\
\bottomrule\end{tabular}\hfill
\begin{tabular}{@{}ll@{}}\toprule Parameter & Value \\ \midrule
Batch size & 128 \\
Learning rate & \(10^{-3}\) \\
Weight decay & 0.01 \\
Optimizer / scheduler & AdamW / cosine \\
Warmup / total epochs & 5 / 50 \\
Min LR / grad clip & \(10^{-6}\) / 1.0 \\
Prediction mode & \(v\)-prediction \\
\bottomrule\end{tabular}
\end{table}

\section{Robot Policy Architecture and Training}
\label{app:robot_setup}

Tiny ablations use a four-layer SigLIP slice and a four-layer Gemma stack. The
full-encoder validation restores all 27 SigLIP layers and the first four
pretrained PaliGemma layers while retaining the same shallow action head. That
head concatenates condition tokens, state tokens, and action--time tokens,
then applies a pre-LayerNorm bidirectional Transformer and projects action
tokens to velocities. Tables~\ref{tab:app_model_arch}
and~\ref{tab:app_train_hparams} specify the reported configurations.

\begin{figure}[!b]
    \centering
    \includegraphics[width=0.75\linewidth]{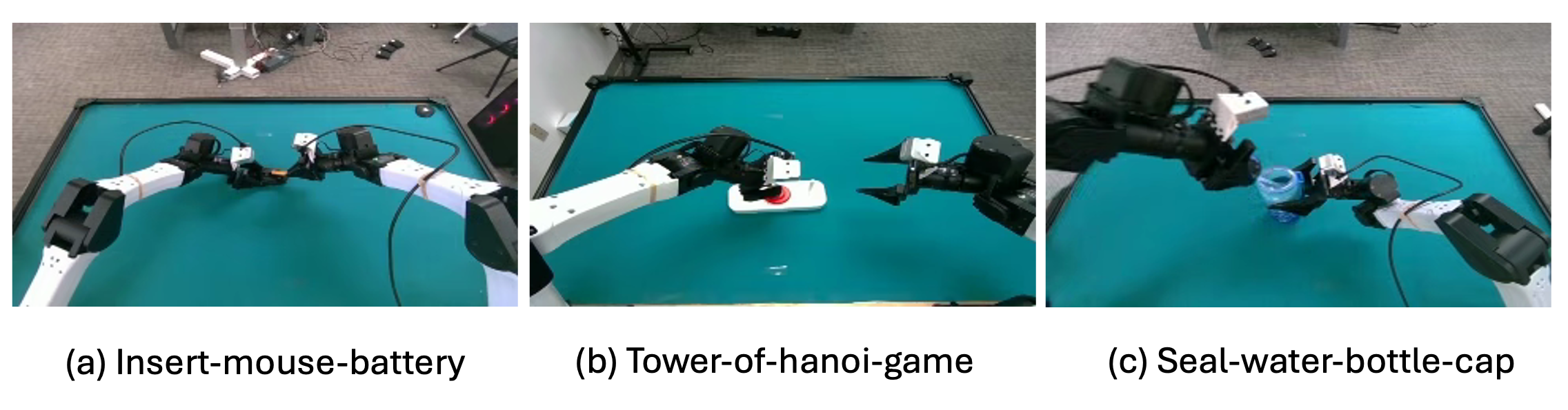}
    \caption{Overview of real-world bimanual manipulation tasks on the DualYam robot platform.}
    \label{fig:yam}
\end{figure}

\begin{table}[h]
\centering
\caption{Parameter breakdown for Tiny and Full policy architectures. Active parameters exclude the frozen vocabulary embedding table.}
\label{tab:param_breakdown}
\small
\begin{tabular}{lrrrr}
\toprule
\textbf{Component} & \multicolumn{2}{c}{\textbf{Tiny (108.7M active)}} & \multicolumn{2}{c}{\textbf{Full (887.6M active)}} \\
\cmidrule(lr){2-3} \cmidrule(lr){4-5}
& Parameters & Size (MB) & Parameters & Size (MB) \\
\midrule
Vision Encoder (SigLIP)       & 62.52M  & 238.5 & 414.80M & 1582.3 \\
Multimodal LLM (Gemma backbone) & 14.95M  & 57.0  & 440.42M & 1679.5 \\
Action Flow Head (+ Proj, Time MLP) & \textbf{31.23M} & 118.3 & \textbf{32.41M} & 123.1 \\
\midrule
Total (incl. Text Embedding)  & 240.40M & 917.1 & 1414.30M & 5395.2 \\
\bottomrule
\end{tabular}
\end{table}

\begin{table}[H]
\centering
\caption{Architecture for the reported LIBERO runs. Horizon controls change only \texttt{action\_horizon}; H10 denotes horizon 10.}
\label{tab:app_model_arch}
\small
\setlength{\tabcolsep}{3.5pt}
\begin{tabular}{@{}llll@{}}
\toprule
Component & Parameter & Tiny & Full encoder \\
\midrule
\multirow{7}{*}{Gemma/VLM}
 & variant & \texttt{gemma\_tiny} & \texttt{gemma\_2b} \\
 & used layers & 4 & 4 \\
 & width & 512 & 2048 \\
 & mlp\_dim & 2048 & 16384 \\
 & num\_heads & 8 & 8 \\
 & num\_kv\_heads & 1 & 1 \\
 & head\_dim & 64 & 256 \\
\midrule
\multirow{2}{*}{Vision}
 & encoder & SigLIP-So400m & SigLIP-So400m \\
 & used layers & 4 & all 27 \\
\midrule
\multirow{4}{*}{Action head}
 & mode / depth & SimVLA / 4 & SimVLA / 4 \\
 & width & 768 & 768 \\
 & heads & 12 & 12 \\
 & mlp\_dim & 3072 & 3072 \\
\midrule
\multirow{3}{*}{Action}
 & action\_dim & 32 & 32 \\
 & action\_loss\_dim & 7 & 7 \\
 & action\_horizon & 10 & 10 \\
\bottomrule
\end{tabular}
\end{table}

\begin{table}[H]
\centering
\caption{Training hyperparameters for the reported LIBERO runs.}
\label{tab:app_train_hparams}
\small
\setlength{\tabcolsep}{4pt}
\begin{tabular}{@{}lll@{}}
\toprule
Parameter & Tiny ablations & Full encoder \\
\midrule
Optimizer & AdamW & AdamW \\
\(\beta_1,\beta_2,\epsilon\) & 0.9, 0.95, \(10^{-8}\) & 0.9, 0.95, \(10^{-8}\) \\
Weight decay / gradient clip & \(10^{-10}\) / 1.0 & \(10^{-10}\) / 1.0 \\
\midrule
LR schedule & cosine decay & cosine decay \\
Warmup / decay steps & 1k / 50k & 1k / 150k \\
Peak / final LR & \(10^{-4}\) / \(10^{-5}\) & \(2{\times}10^{-4}\) / \(2{\times}10^{-5}\) \\
Backbone LR multiplier & 0.1 & 0.1 \\
EMA decay & 0.99 & 0.99 \\
Batch size / steps & 64 / 50k & 256 / 150k \\
\bottomrule
\end{tabular}
\end{table}

\section{Evaluation Protocol and Controls}
\label{app:eval_controls}

A policy query produces an action chunk of horizon \(H\). We separately report
the number of Euler solver steps (\texttt{num\_flow\_steps}, abbreviated
\(fs\)) and the number of environment actions executed before the next policy
query (\texttt{replan\_steps}, abbreviated \(r\)). Unless stated otherwise,
the main results use H10 and execute the entire predicted chunk (\(r=10\)).
This distinction matters: \(fs=1\) specifies one flow evaluation per query,
not an open-loop rollout length.

\begin{figure}[!htbp]
\centering
\includegraphics[width=0.72\linewidth]{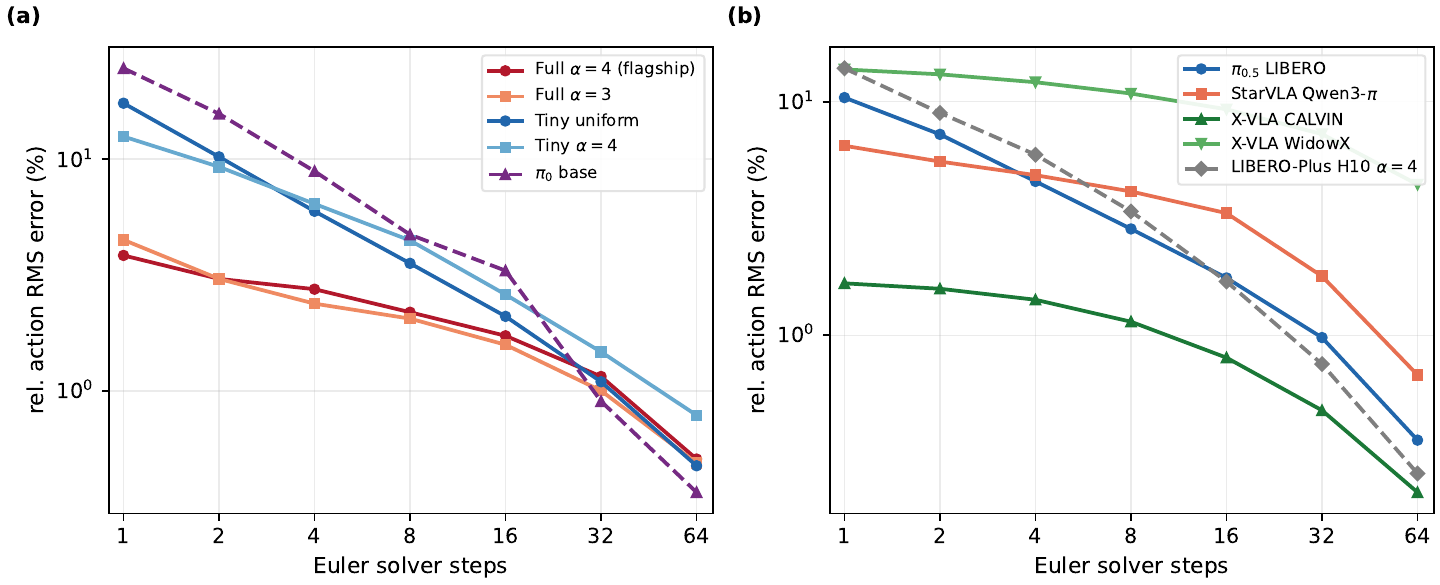}
\caption{Solver-step convergence audit. Relative action RMS error of Euler integration with $fs\in\{1,2,4,8,16,32,64\}$ steps against a 128-step Euler reference, measured on fixed observations. (a) Our LIBERO checkpoints; (b) third-party checkpoints and a LIBERO-Plus recipe. The error drops below $1\%$ by 64 steps for all but X-VLA WidowX, while the one-step error remains small ($1.7$--$19\%$), consistent with the strong-condition signature of Fig.~\ref{fig:vloss_profiles}.}
\label{fig:solver_audit}
\end{figure}

Table~\ref{tab:app_replan_h10} gives the H10 replanning sensitivity; all
comparisons keep \(fs\), \(r\), and horizon separate.

\begin{table}[H]
\centering
\caption{H10 replanning sensitivity on LIBERO-Long. Values are success rates (\%).}
\label{tab:app_replan_h10}
\footnotesize
\setlength{\tabcolsep}{4pt}
\begin{tabular}{llcccc}
\toprule
Recipe & Flow steps & \(r=1\) & \(r=2\) & \(r=5\) & \(r=10\) \\
\midrule
Uniform & 1 & 32.0 & 38.4 & 72.8 & 70.2 \\
Uniform & 10 & 55.0 & 58.0 & 80.2 & 80.8 \\
Noise shift (\(\alpha=3\)) & 1 & 24.0 & 33.2 & 79.2 & 78.0 \\
Noise shift (\(\alpha=3\)) & 10 & 15.6 & 25.4 & 52.4 & 52.4 \\
Noise shift (\(\alpha=4\)) & 1 & 33.0 & 45.0 & 86.2 & 85.2 \\
Noise shift (\(\alpha=4\)) & 10 & 18.0 & 29.0 & 66.8 & 63.4 \\
Noise shift (\(\alpha=8\)) & 1 & 29.4 & 39.8 & 81.4 & 78.0 \\
Noise shift (\(\alpha=8\)) & 10 & 3.4 & 7.0 & 22.6 & 22.6 \\
Pure noise & 1 & 23.2 & 33.2 & 73.6 & 72.8 \\
\bottomrule
\end{tabular}
\end{table}

\subsection{Inference Latency Decomposition}
\label{app:latency}

\begin{figure}[!htbp]
\centering
\includegraphics[width=0.85\linewidth]{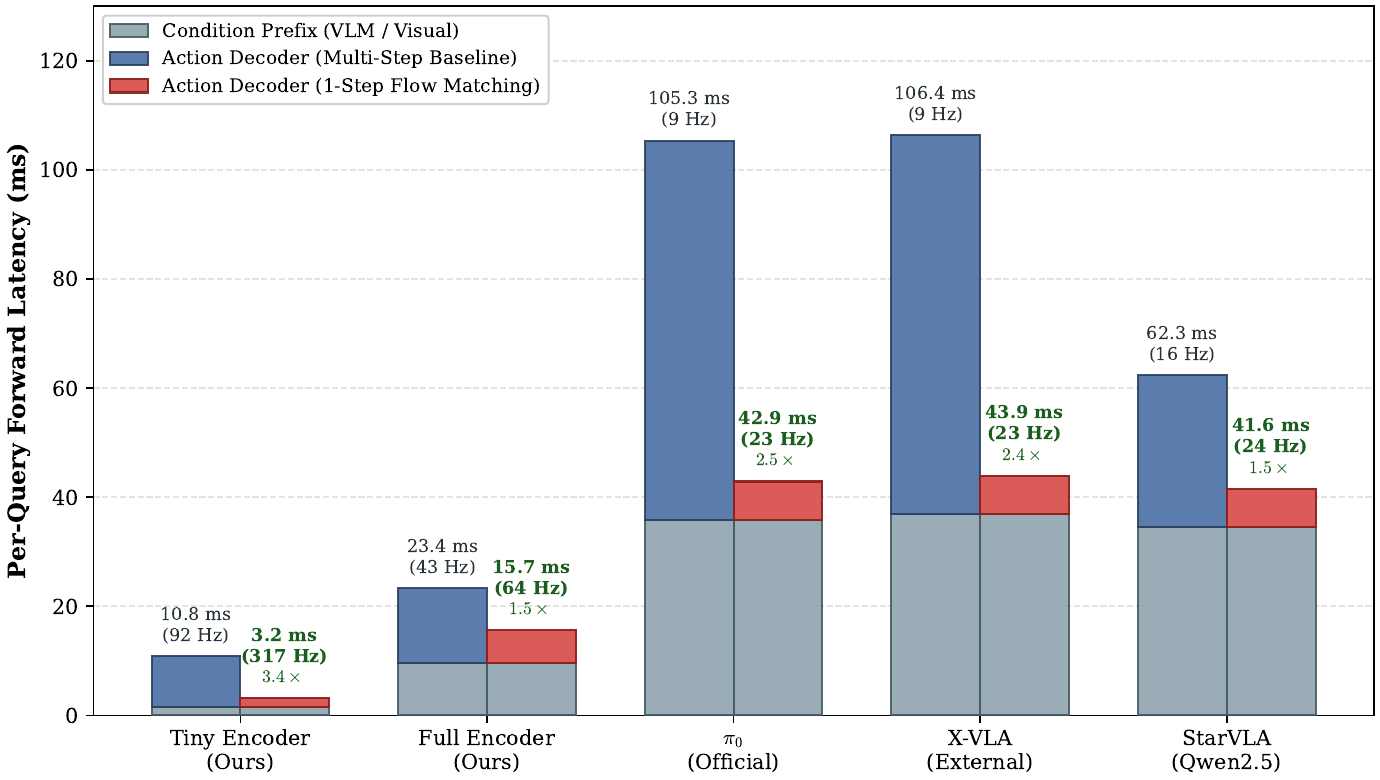}
\caption{\textbf{Inference latency decomposition} (RTX 5090, batch 1, model forward). Condition prefix vs.\ action decoder at the multi-step baseline (10-step; 4-step for StarVLA) and at one-step decoding. Single-stepping cuts total per-query latency \(1.5\)--\(3.4\times\), after which the condition prefix dominates the remainder.}
\label{fig:latency_decomposition}
\vspace{-0.6em}
\end{figure}

Figure~\ref{fig:latency_decomposition} decomposes per-query model-forward latency (NVIDIA RTX 5090, batch size 1, mean of 50 warmed calls on synthetic observations with the deployed token counts; excludes input/output transforms, networking, and the simulator) into the condition prefix and the action decoder, at each architecture's multi-step baseline and at one-step decoding.
Single-stepping cuts total latency by \(1.5\)--\(3.4\times\) across all five architectures.
The condition prefix then dominates the remainder for our policies (52\% and 62\% of the Tiny and Full forwards) and accounts for 83\% of the official \(\pi_0\) forward---the systems-level counterpart of the condition-dominance picture: once decoding is single-step, the remaining cost is condition construction, not action generation.
Our two columns are profiled from the deployed JAX path (checkpoints of Tables~\ref{tab:time_ablation} and~\ref{tab:full_encoder}); the \(\pi_0\) column is profiled on the official DROID checkpoint, which shares its architecture and horizon with the LIBERO one; the X-VLA and StarVLA columns reuse our earlier profiling under the same protocol.

\FloatBarrier

\section{LIBERO-Plus Full-Condition Sweep}
\label{app:libero_plus}

\begin{figure}[H]
\centering{\includegraphics[height=3.7cm]{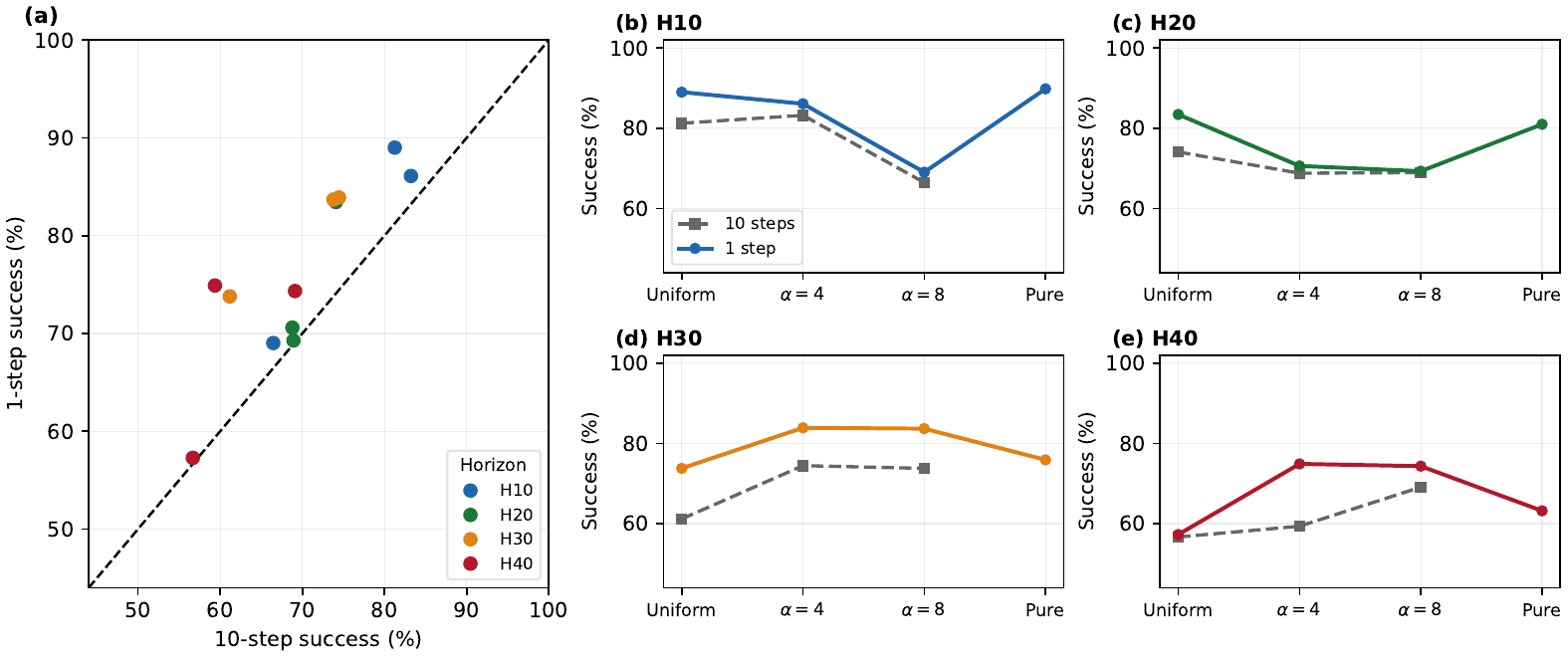}}%
\hfill
{\includegraphics[height=3.7cm]{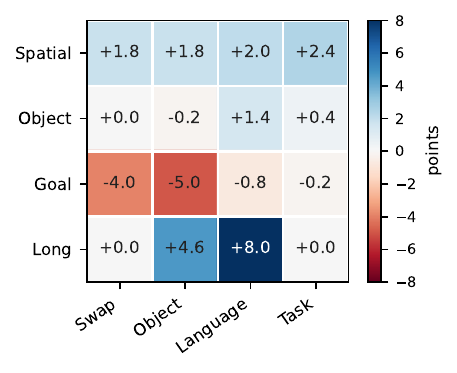}}
\caption{\textbf{One-step parity under distribution shift.} \textbf{Left:} LIBERO-Plus full-condition sweep across horizons and schedules (panel labels (a)--(e) are inside this panel); all 18 comparable non-pure recipes lie on or above the diagonal (+5.4 pp mean margin). \textbf{Right:} LIBERO-Pro robustness probe, showing the success-rate difference between 1-step and 10-step decoding across the 16 perturbation cells.}
\label{fig:shift_parity}
\end{figure}

We evaluate the tiny policy directly on LIBERO-Plus~\citep{fei2025liberoplus} across horizons and noise schedules (~2500 episodes/suite). As shown in the left panel of Fig.~\ref{fig:shift_parity}, across 18 comparable non-pure recipes, one-step decoding consistently matches or exceeds ten-step decoding (+5.4 pp mean margin), confirming that the high-noise one-step benefit generalizes under distribution shifts.

\paragraph{LIBERO-Pro robustness probe.}
We test the full-encoder standard-LIBERO checkpoint directly on LIBERO-Pro perturbations~\citep{zhou2026liberopro} without task-specific tuning (500 episodes/cell, right panel of Fig.~\ref{fig:shift_parity}). One-step decoding averages 44.2\% across all cells, while ten-step averages 43.5\% (15 of 16 cells within 5 points; per-perturbation one-step means: Swap 2.6, Object 75.5, Language 82.6, Task 16.2), confirming that one-step decoding is robust to perturbations without sacrificing generalization.

Table~\ref{tab:app_libero_plus_fs1} provides the exact aggregate values. Each
recipe is evaluated across the four suites with around 2,500 episodes per
suite. Values are means over the four suites; blank cells correspond to
recipes not trained in the sweep.

\begin{table}[H]
\centering
\caption{LIBERO-Plus full-condition one-step sweep. \(U\) denotes uniform; numbered columns denote \(\alpha\). Values are four-suite mean success rates (\%).}
\label{tab:app_libero_plus_fs1}
\scriptsize
\renewcommand{\arraystretch}{0.85}
\setlength{\tabcolsep}{4pt}
\begin{tabular}{lccccccc}
\toprule
Horizon & U & 1 & 2 & 3 & 4 & 8 & Pure \\
\midrule
H10 & 89.0 & 92.9 & 94.3 & 80.7 & 86.1 & 69.0 & 89.8 \\
H20 & 83.4 & 68.2 & 49.1 & 63.6 & 70.6 & 69.3 & 81.0 \\
H30 & 73.8 & -- & -- & -- & 83.9 & 83.7 & 75.9 \\
H40 & 57.3 & -- & -- & -- & 74.9 & 74.3 & 63.2 \\
\bottomrule
\end{tabular}
\end{table}

\FloatBarrier

\section{Action-Head Positional Controls}
\label{app:decoder_controls}

The deeper-head ablations use the same tiny-model setup as the main LIBERO
controls; the per-layer position variant adds a learned absolute position
embedding inside each action-decoder block. These are preliminary probes: every variant is trained once under the shared
hyperparameters of Table~\ref{tab:app_train_hparams} with no per-variant
tuning.
Our preliminary experiments suggest that scaling the action head brings no clear benefit.

\begin{table}[H]
\centering
\caption{Depth-8 action-head controls (1-step / 10-step success \%). Top: per-suite under H10 \(\alpha{=}4\). Bottom: LIBERO-Long vs.\ train-time \(\alpha\).}
\label{tab:app_depth8}\label{tab:app_depth8_long}
\scriptsize
\renewcommand{\arraystretch}{0.85}
\setlength{\tabcolsep}{4pt}
\begin{tabular}{lccccc}
\toprule
Variant & Train \(\alpha\) & Spatial & Object & Goal & Long \\
\midrule
Vanilla depth8 & 4 & 95.6\,/\,70.0 & 99.0\,/\,53.6 & 97.0\,/\,69.6 & 80.2\,/\,2.6 \\
Per-layer position & 4 & 97.4\,/\,84.6 & 98.8\,/\,83.2 & 95.4\,/\,73.4 & 60.2\,/\,9.6 \\
RoPE & 4 & 97.0\,/\,69.6 & 99.2\,/\,57.8 & 94.4\,/\,65.8 & 80.0\,/\,8.8 \\
\midrule
Shallow baseline & 4 & -- & -- & -- & 85.2\,/\,63.4 \\
Depth8 RoPE & 3 & -- & -- & -- & 74.6\,/\,41.2 \\
Depth8 RoPE & 8 & -- & -- & -- & 70.2\,/\,25.8 \\
\bottomrule
\end{tabular}
\end{table}

\section{Trajectory Dynamics of the Deployed One-Step Map}
\label{app:traj_dynamics}

We evaluate the deployed one-step map $T_c(z)=z+v_\theta(z,t{=}0;c)$ ($\alpha{=}4$) frame-by-frame across four held-out LIBERO demonstrations totaling 855 observations (episodes 10, 392, 0, and 380). At each frame, we decode an $H$-step action chunk across $n_z=8$ fixed Gaussian noise sources, form its $7H\times 7H$ Jacobian, and compute the exact leading singular value $\sigma_1$ via SVD for four separately trained horizon checkpoints ($H \in \{10,20,30,40\}$).

\begin{figure}[!htbp]
\centering
\includegraphics[width=\linewidth]{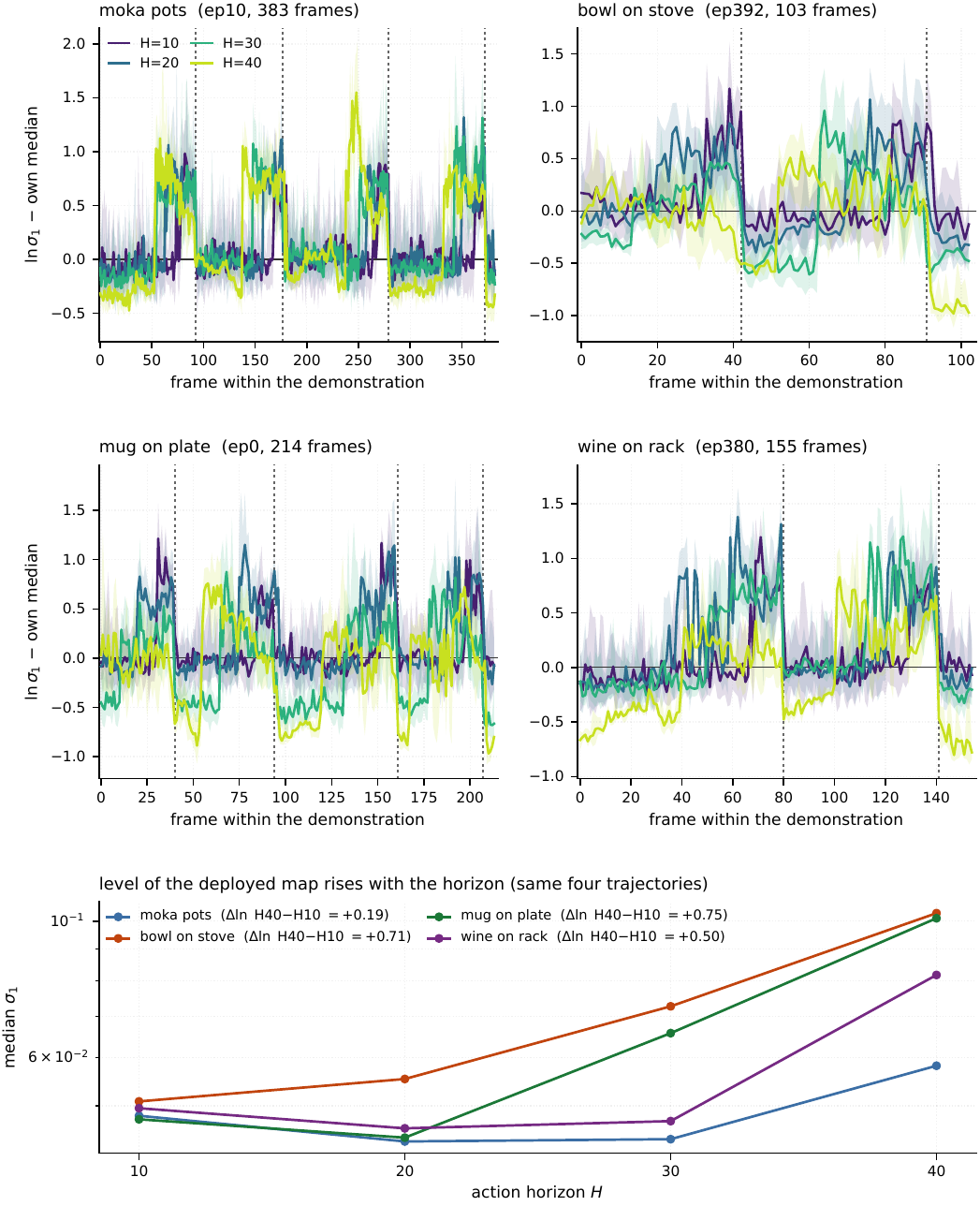}
\caption{\textbf{Frame-by-frame Jacobian spectral norm ($\sigma_1$) across four LIBERO demonstrations.} \textbf{Top:} Median-centered $\ln\sigma_1$ (mean $\pm$ min--max range over $n_z=8$ noise sources) for separately trained checkpoints ($H \in \{10,20,30,40\}$), with dotted verticals marking ground-truth gripper switches. \textbf{Bottom:} Unshifted trajectory medians across horizons.}
\label{fig:app_traj_dynamics}
\vspace{-0.6em}
\end{figure}

As shown in Figure~\ref{fig:app_traj_dynamics}, two findings emerge:
\begin{enumerate}
    \item \textbf{Sensitivity spikes align with physical transitions:} The spectral excursions analyzed in Appendix~\ref{app:onestep_geometry} are not diffuse noise; across all four demonstrations, $\sigma_1$ peaks lock onto gripper switch boundaries, showing that the map's sharpest directions occur precisely at physical decision points.
    \item \textbf{Baseline sensitivity scales with horizon:} Unshifted median $\sigma_1$ at $H{=}40$ exceeds that at $H{=}10$ in every demonstration by $\Delta\ln\sigma_1 \in [0.19, 0.76]$ ($1.2\times$--$2.1\times$), matching the trend in Fig.~\ref{fig:horizon_vloss}. Because each horizon uses a separately trained checkpoint, elevated sensitivity is an intrinsic property of long-horizon models rather than an artifact of rolling out a single model further.
\end{enumerate}

\end{document}